\documentclass{article}

    \usepackage[preprint]{neurips_2025}

\usepackage{microtype}
\usepackage{hyperref}
\usepackage{url}
\usepackage{booktabs}
\usepackage{times}
\usepackage{latexsym}
\usepackage{amsmath}
\usepackage{amsfonts}
\usepackage{graphicx}
\usepackage{multirow}
\usepackage{subcaption}
\usepackage[T1]{fontenc}

\usepackage{tikz}

\usepackage{listings}
\usepackage{color}
\definecolor{dkgreen}{rgb}{0,0.6,0}
\definecolor{gray}{rgb}{0.5,0.5,0.5}
\definecolor{mauve}{rgb}{0.58,0,0.82}
\definecolor{lightblue}{rgb}{0.9, 0.95, 1.0}
\definecolor{lightgray}{rgb}{0.9, 0.9, 0.9}

\usepackage{tikz}

\usepackage{colortbl}
\usepackage{pifont}
\usepackage{algorithmic}
\usepackage{algorithm}

\usepackage[utf8]{inputenc}
\usepackage{multirow}
\usepackage{microtype}
\usepackage{booktabs}

\usepackage{wrapfig}    

\usepackage{hyperref}
\usepackage{url}

\usepackage{pifont}
\usepackage{microtype}
\usepackage{hyperref}
\usepackage{url}
\usepackage{booktabs}
\usepackage{times}
\usepackage{latexsym}
\usepackage{amsmath}
\usepackage{amsfonts}
\usepackage{graphicx}
\usepackage{multirow}
\usepackage{microtype}
\usepackage{booktabs}
\usepackage{graphicx}
\usepackage{subcaption}
\usepackage{wrapfig}
\usepackage{enumitem}

\usepackage{CJKutf8}
\newcommand{\zh}[1]{\begin{CJK}{UTF8}{gbsn}#1\end{CJK}}

\definecolor{-}{rgb}{0.70,0.13,0.13}
\definecolor{+}{rgb}{0.0, 0.6, 0.3}

\title{The Emergence of Abstract Thought in Large Language Models Beyond Any Language}

\author{
  \hspace{-0.42cm}Yuxin Chen$^{1}$\footnotemark[1] \quad Yiran Zhao$^{2}$\footnotemark[1] \quad Yang Zhang$^{3}$ \quad An Zhang$^{1}$  \quad Kenji Kawaguchi$^{1}$ \quad Shafiq Joty$^{2}$ \\
  \textbf{Junnan Li}$^{2}$ \quad \textbf{Tat-Seng Chua}$^{1}$ \quad \textbf{Michael Qizhe Shish}$^{1}$\footnotemark[2] \quad \textbf{Wenxuan Zhang}$^{4}$\footnotemark[2] \\
   $^1$ National University of Singapore \quad $^2$ Salesforce AI Research \quad  $^3$ Peking University \\
  \hspace{-0.42cm}$^4$ Singapore University of Technology and Design \\
}

\begin{document}

\maketitle
\renewcommand{\thefootnote}{\fnsymbol{footnote}}
\footnotetext[1]{Equal Contribution.}
\footnotetext[2]{Corresponding Authors.}
\renewcommand{\thefootnote}{\arabic{footnote}}

\begin{abstract}

As large language models (LLMs) continue to advance, their capacity to function effectively across a diverse range of languages has shown marked improvement. Preliminary studies observe that the hidden activations of LLMs often resemble English, even when responding to non-English prompts. This has led to the widespread assumption that LLMs may ``think'' in English. However, more recent results showing strong multilingual performance, even surpassing English performance on specific tasks in other languages, challenge this view.
In this work, we find that LLMs progressively develop a core language-agnostic parameter space—a remarkably small subset of parameters whose deactivation results in significant performance degradation across all languages. This compact yet critical set of parameters underlies the model’s ability to generalize beyond individual languages, supporting the emergence of abstract thought that is not tied to any specific linguistic system.
Specifically, we identify language-related neurons—those are consistently activated during the processing of particular languages, and categorize them as either shared (active across multiple languages) or exclusive (specific to one). As LLMs undergo continued development over time, we observe a marked increase in both the proportion and functional importance of shared neurons, while exclusive neurons progressively diminish in influence. These shared neurons constitute the backbone of the core language-agnostic parameter space, supporting the emergence of abstract thought. 
Motivated by these insights, we propose neuron-specific training strategies tailored to LLMs' language-agnostic levels at different development stages. Experiments across diverse LLM families support our approach.\footnote{Our codes are available at \url{https://github.com/chenyuxin1999/Abstract_Thought}.}

\end{abstract}


\section{Introduction}

As large language models (LLMs) continue to advance~\citep{openai2023gpt4, touvron2023llama, hurst2024gpt, yang2024qwen2technicalreport, team2024gemini}, their performance across a wide range of languages (known as multilingual capability) has markedly improved over the past years~\citep{le2023bloom, yang2024qwen2, ustun2024aya}. 
Despite this progress, several studies have observed that LLMs tend to ``think in English'', often using it as an internal language of thought even when processing inputs in other languages~\citep{wendler2024llamas, how_do_handle, schut2025multilingual}. 
This phenomenon has led to the hypothesis that LLM performance in non-English languages is inherently constrained by their capabilities in English~\citep{qin2023cross, liu2024translation}.

Yet, more recent findings complicate this narrative: some studies found that LLMs can actually outperform their English-language performance on certain tasks in other languages~\citep{zhao2025babel, team2024gemma2, team2025gemma}, indicating that non-English processing may not always rely on English as an intermediate language. These conflicting observations raise a deeper research question: \textit{Do LLMs think in the distinct space of each language}, or \textit{Do they operate in a higher-level language-agnostic space beyond any specific language?}
In other words, whether the trend of non-English performance compared to English indicates \textbf{the emergence of abstract thought within LLMs?}

\begin{wrapfigure}{r}{0.44\textwidth}
    \vspace{-0.2cm}
    \centering
    \includegraphics[width=0.44\textwidth]{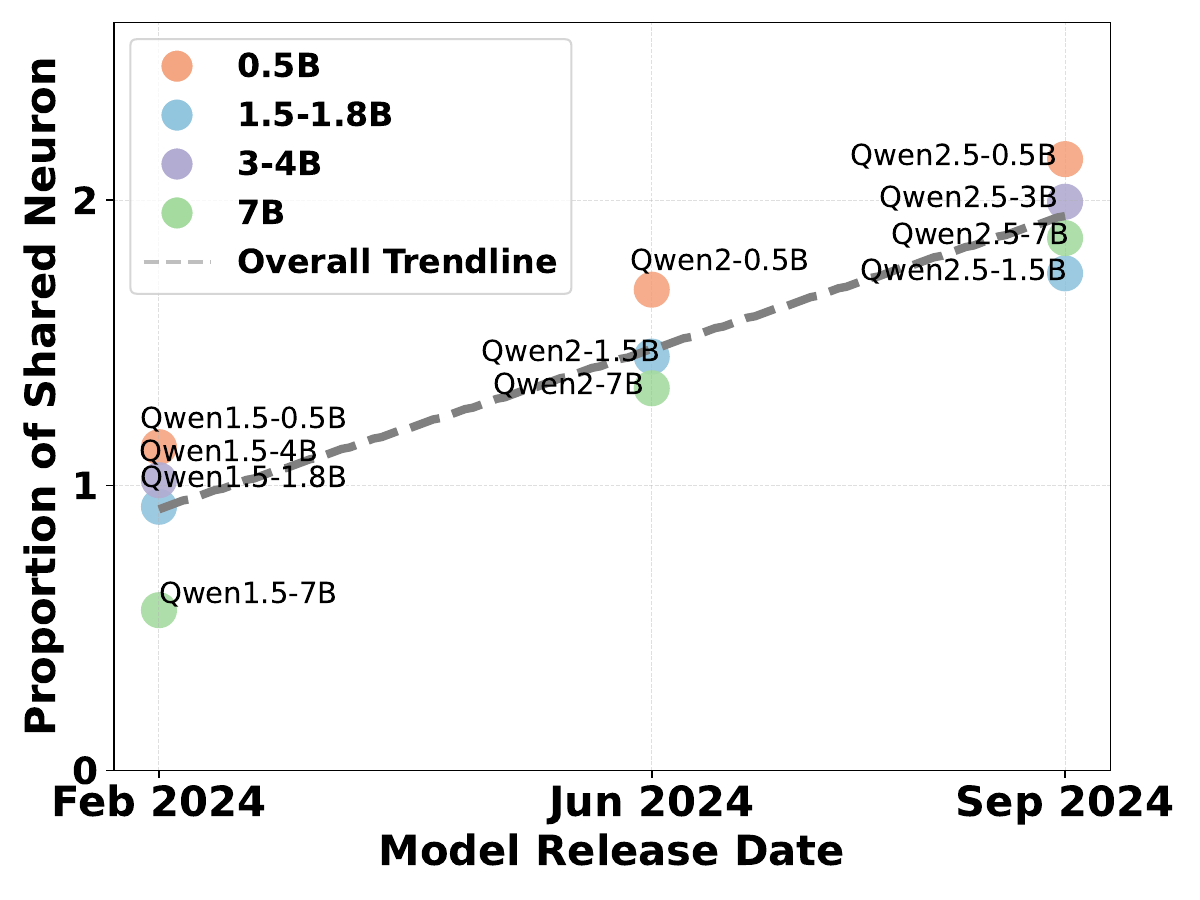}
    \vspace{-0.2cm}
    \includegraphics[width=0.44\textwidth]{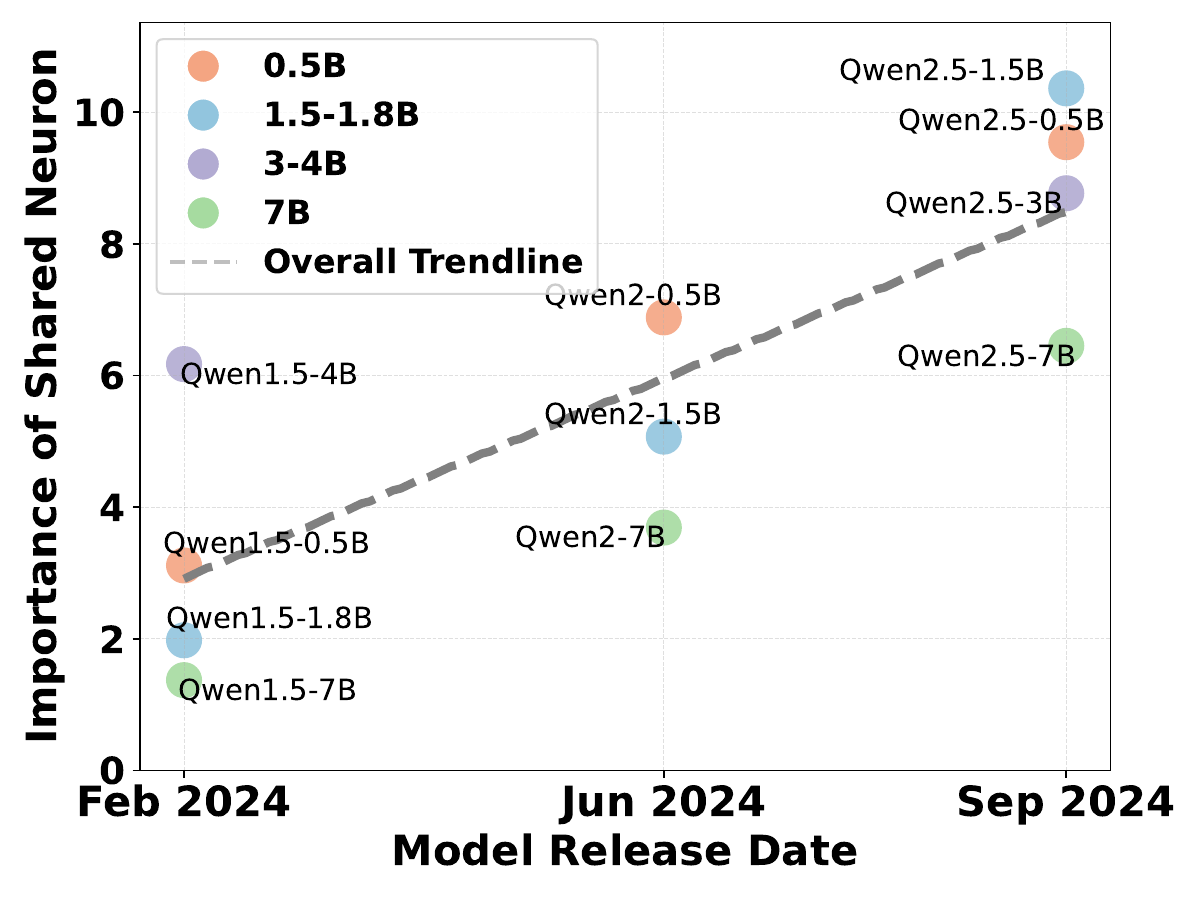}
    \vspace{-0.2cm}
    
    \caption{
    \textbf{(Top)} The trendline of shared neuron proportion rises with model release date (see Section~\ref{shared neuron ratio}). 
    \textbf{(Bottom)} The trendline of shared neuron importance also grows, indicating their increasing language-agnostic property (see Section~\ref{shared neuron importance}).
    }
    \label{fig:scaling}
    \vspace{-0.2cm}
\end{wrapfigure}

In this work, we explore the existence and development of abstract thought in LLMs by analyzing how individual neurons, responsible for models' thinking, respond to multilingual queries. Here each neuron corresponds to a row or column in the model's parameter matrices and is considered activated if its removal significantly alters the model's output \citep{frankle2018lottery, tang2024language, wang2025investigating}. We begin by identifying neurons activated when the model processes inputs in specific languages, referred to as \textit{Language-Related Neurons}. 
To investigate whether these neurons become increasingly specialized for specific languages or potentially exhibit more general and language-agnostic functionality, we distinguish between \textit{Language-Exclusive Neurons}, which are activated only for one language, and \textit{Language-Shared Neurons}, which are consistently activated across all languages considered.
Figure~\ref{fig:scaling} (top) shows a positive correlation between multilingual ability and the proportion of language-shared neurons across different generations of LLMs. This suggests that \textbf{as multilingual performance improves, a greater proportion of language-related neurons are shared across languages.}

Building on the observed positive relationship between language-shared neurons and multilingual capability, as well as the finding that LLMs increasingly outperform in non-English languages on certain tasks, we hypothesize that shared neurons may gradually assume more fundamental roles beyond merely supporting multilingual processing. 
Accordingly, rather than focusing solely on how the proportion of language-shared neurons evolves across model generations, it is essential to evaluate their functional significance relative to language-exclusive neurons. If shared neurons contribute more critically to multilingual processing than language-exclusive neurons—which also participate in language tasks and should be comparably important in principle—this would indicate that shared neurons have evolved into \textit{Language-Agnostic Neurons}, which go beyond shared activation patterns to support abstract functions like semantic reasoning and generalization. As these neurons evolve, they support increasingly abstract thought that transcends linguistic boundaries. As shown in Figure \ref{fig:scaling} (bottom), language-shared neurons exhibit a markedly growing importance in multilingual processing relative to language-exclusive neurons, signaling \textbf{the emergence of language-agnostic properties and potentially, the development of abstract thought in LLMs.}

Inspired by the insights discussed above, we propose a set of targeted neuron training strategies aimed at enhancing the multilingual capabilities of LLMs. These methods are tailored based on the presence or absence of language-agnostic neurons, which serve as an indicator of the emergence of abstract thought within the model. For LLMs that lack language-agnostic neurons, the model is likely still under-trained; thus, training any language-related neurons can contribute to improving multilingual performance. In contrast, in LLMs where abstract thought has emerged, language-shared neurons have evolved into language-agnostic ones. As these neurons have reached a form of generalization, further improvement through additional training is limited. In such cases, enhancing multilingual capabilities requires focusing on training language-exclusive neurons to better support language-specific nuances. We validate our approach through comprehensive experiments across diverse model series and release time. The results demonstrate that our training method, guided by the presence of language-agnostic properties, effectively enhances multilingual performance.

\section{Metrics for Exploring Abstract Thought}

In this section, we identify neurons associated with language processing, referred to as \textit{Language-Related Neurons}, and, based on them, define several metrics to quantify and analyze the emergence of abstract thought in LLMs.

\subsection{Language-Related Neurons}

We identify language-related neurons as those that are consistently activated when processing inputs in a particular language, where a neuron is defined as a single row or column within the model's parameter matrices. Building on prior work in identifying important neurons in neural networks~\citep{frankle2018lottery, ni2023finding, tang2024language, how_do_handle}, we consider a neuron to be activated if its removal leads to a significant change in the resulting embedding. Formally, given an input sequence $x$ in a specific language, a neuron $\mathcal{N}$ is considered activated if
\begin{equation}
\left\| \mathcal{LLM}(x) - \mathcal{LLM}_{\ominus \mathcal{N}}(x) \right\|_2 \geq \sigma,
\end{equation}
where $\mathcal{LLM}(x)$ denotes the output embedding when processing $x$, and $\mathcal{LLM}_{\ominus \mathcal{N}}(x)$ denotes the output when neuron $\mathcal{N}$ is deactivated, i.e., its parameters are set to zero. The threshold $\sigma$ specifies the minimum magnitude of change required to consider a neuron activated. 

Furthermore, language-related neurons $\mathcal{N}^{\ell}_{\mathrm{lang}}$ for a specific language $\ell$ are identified through 
\begin{equation}
\mathcal{N}^{\ell}_{\mathrm{lang}} := \Big\{ \mathcal{N} \in \mathcal{LLM}\;\Big|\;\left\| \mathcal{LLM}(x) - \mathcal{LLM}_{\ominus \mathcal{N}}(x) \right\|_2 \geq \sigma,\; \forall x\in \ell \Big\}.\label{equ:neuron}
\end{equation}

Since sequentially deactivating neurons in Equation \ref{equ:neuron} is computationally expensive, we employ the parallel neuron detection methods proposed in \citet{how_do_handle, wang2025investigating}. Further implementation details are provided in Appendix~\ref{sec:parallel}.

\subsection{Language-Shared and Language-Exclusive Neurons}
\label{shared neuron ratio}

To investigate whether neurons become increasingly specialized for specific languages or exhibit language-agnostic behavior, we conduct a preliminary analysis of the proportion of \textit{Language-Shared Neurons}, defined as language-related neurons that are consistently activated across all languages considered, and \textit{Language-Exclusive Neurons}, defined as language-related neurons that are uniquely activated for individual languages and not shared across all languages. Formally, language-shared and language-exclusive neurons are defined as follows:
\begin{equation}
\mathcal{N}_{\mathrm{shared}} := \bigcap_{\ell \in \mathcal{L}} \mathcal{N}_{\mathrm{lang}}^{\ell}, \quad \text{and} \quad \mathcal{N}_{\mathrm{exclusive}}^{\ell} := \mathcal{N}_{\mathrm{lang}}^{\ell} \setminus \mathcal{N}_{\mathrm{shared}},
\end{equation}
where $\mathcal{L}$ denotes the set of all languages under consideration. In other word, $\mathcal{N}_{\mathrm{shared}}$ consistently exhibit high importance across inputs from different languages, while $\mathcal{N}_{\mathrm{exclusive}}^{\ell}$ is the set of neurons specific to that language but not part of the shared set. Furthermore, we examine the proportion of language-shared neurons relative to language-exclusive neurons, defined as
\begin{equation}
\mathrm{Shared\;Neuron\;Ratio} := \frac{|\mathcal{N}_{\mathrm{shared}}|}{\frac{1}{|\mathcal{L}|} \sum_{\ell \in \mathcal{L}} |\mathcal{N}_{\mathrm{exclusive}}^{\ell}|},\label{equ:share}
\end{equation}
which quantifies the extent to which individual neurons are shared across all languages as opposed to being specialized for specific ones. A higher ratio indicates a greater number of neurons that are commonly activated across languages, while a lower ratio suggests that more neurons are uniquely responsive to individual languages.

\subsection{Language-Agnostic Neurons}
\label{shared neuron importance}

As LLMs continue to improve in their ability to handle multiple languages, and even outperform their English capabilities on specific tasks, we hypothesize that shared neurons may gradually serve more fundamental functions beyond the processing of multiple languages. Consequently, rather than solely examining the evolution of the proportion of language-shared neurons across successive model generations, it is also crucial to assess their relative functional significance in comparison to language-exclusive neurons. 
Specifically, if language-shared neurons contribute significantly more to multilingual processing than language-exclusive neurons, this indicates a functional difference between the two, since both types are involved in language-related tasks and identified using the same criteria, they should be equally important if their roles are analogous. The discrepancy suggests that shared neurons have evolved into \textit{Language-Agnostic Neurons}.
Note that while language-shared neurons are activated across multiple languages, language-agnostic neurons reflect a higher level of abstraction. Rather than encoding language-specific features, they are hypothesized to support cognitive functions that transcend individual languages, such as semantic abstraction, reasoning, and generalization.

To investigate whether language-agnostic property emerge in language-shared neurons, we introduce the metric \textit{Language-Shared Neuron Importance}, which quantifies the impact of deactivating language-shared neurons versus language-exclusive neurons on the model's performance in a given language. This is operationalized by measuring the change in perplexity (\(\Delta \mathrm{PPL}\)) when each neuron group is ablated. A disproportionately larger increase in perplexity upon deactivating shared neurons would suggest their greater functional importance.
Formally, we define the language-shared neuron importance for a language $\ell$ as:
\begin{equation}
    \mathrm{Imp}^{\ell} := \frac{\Delta \mathrm{PPL}_{\mathrm{shared}}^{\ell} / |\mathcal{N}_{\mathrm{shared}}|}{\Delta \mathrm{PPL}_{\mathrm{exclusive}}^{\ell} / |\mathcal{N}_{\mathrm{exclusive}}^{\ell}|},
\end{equation}
where \(\Delta \mathrm{PPL}_{\mathrm{shared}}^{\ell}\) and \(\Delta \mathrm{PPL}_{\mathrm{exclusive}}^{\ell}\) denote the changes in perplexity for language \(\ell\) when shared and corresponding exclusive neurons are deactivated, respectively, and \(|\mathcal{N}_{\mathrm{shared}}|\) and \(|\mathcal{N}_{\mathrm{exclusive}}^{\ell}|\) represent the number of neurons in each group. A higher value of $\mathrm{Imp}^{\ell}$ indicates that shared neurons contribute more significantly than language-exclusive neurons, thereby providing evidence for their language-agnostic role, since both types of neurons should exhibit comparable importance if their functions were equivalent.

To obtain an overall model-level estimation reflecting this trend across different languages,
we compute the average importance across all languages and apply a logarithmic transformation to mitigate scale sensitivity. We refer to the resulting quantity as the \textit{Language Agnostic Score}:
\begin{equation}\
\begin{aligned}
\mathrm{Language}\;& \mathrm{Agnostic\;Score} :=  \log\left(1 + \frac{1}{|\mathcal{L}|} \sum_{\ell \in \mathcal{L}} \mathrm{Imp}^{\ell} \right),\label{equ:agnostic}
\end{aligned}
\end{equation}
which quantifies the average degree to which language-shared neurons contribute in a language-agnostic manner across the evaluated languages. In contrast to the shared neuron ratio defined in Equation \ref{equ:share}, which solely quantifies the number of shared neurons, the language-agnostic score incorporates the functional importance of neurons. Higher values suggest not only stronger language-agnostic behavior but also hint at the emergence of abstract thought in LLMs.

\section{Emergence of Abstract Thought}

\subsection{Experiment Setup}
\label{subsec:neuron_setup}

\paragraph{Evaluated Models}

To comprehensively evaluate the emergence of abstract thought in LLMs throughout their development, we examine 20 open-source models encompassing diverse model families, release periods, and sizes. Specifically, we evaluate Llama series including LLaMA1-7B~\citep{touvron2023llama}, Llama2-7B~\citep{touvron2023llama2}, Llama3.2-1B, Llama3.2-3B, Llama3-8B, Llama3.1-8B~\citep{grattafiori2024llama}, Qwen1.5-0.5B, Qwen1.5-1.8B, Qwen1.5-4B, Qwen1.5-7B~\citep{bai2023qwen}, Qwen2-0.5B, Qwen2-1.5B, Qwen2-7B~\citep{yang2024qwen2technicalreport}, Qwen2.5-0.5B, Qwen2.5-1.5B, Qwen2.5-3B, Qwen2.5-7B~\citep{yang2024qwen2}, Gemma-7B~\citep{team2024gemma}, Gemma2-9B~\citep{team2024gemma2}, Gemma3-4B~\citep{gemma3}.

\paragraph{Multilingual Benchmark} 

We evaluate models across six typologically and resource-diverse languages: Chinese (Zh), English (En), Thai (Th), Swahili (Sw), French (Fr), and German (De). This selection spans high-resource, medium-resource, and low-resource languages, enabling a representative analysis of language-related neuron behaviors. For our analysis, we utilize the Multilingual Massive Multitask Language Understanding (MMMLU) dataset~\citep{openai2024mmmlu}, a human-translated extension of the original MMLU benchmark~\citep{hendrycks2021measuringmassivemultitasklanguage}, available in 14 languages. In addition, we incorporate the Multilingual Grade School Math (MGSM) dataset~\citep{shi2022language}, a translated version of GSM8K~\citep{cobbe2021gsm8k}, which covers 10 languages. Together, these datasets provide quantitative measures of the models’ multilingual capabilities.

\vspace{-2mm}
\paragraph{Neuron Detection Corpus} For each language, we identify language-related neurons by analyzing activation patterns on $1000$ sentences sampled from the OSCAR corpus \citep{oscar}. To quantify the functional contribution of these neurons, we further compute perplexity changes caused by deactivating them, using the same language-specific OSCAR data. This unified framework allows us to assess both the proportion and the importance of language-specific and shared neurons across languages and model generations. More detailed illustration can be found in Appendix \ref{sec:corpus}.

\subsection{Analysis on Shared Neuron Ratio}

\begin{figure}[t]
    \centering
    \vspace{-5pt}
    \includegraphics[width=1.0\textwidth]{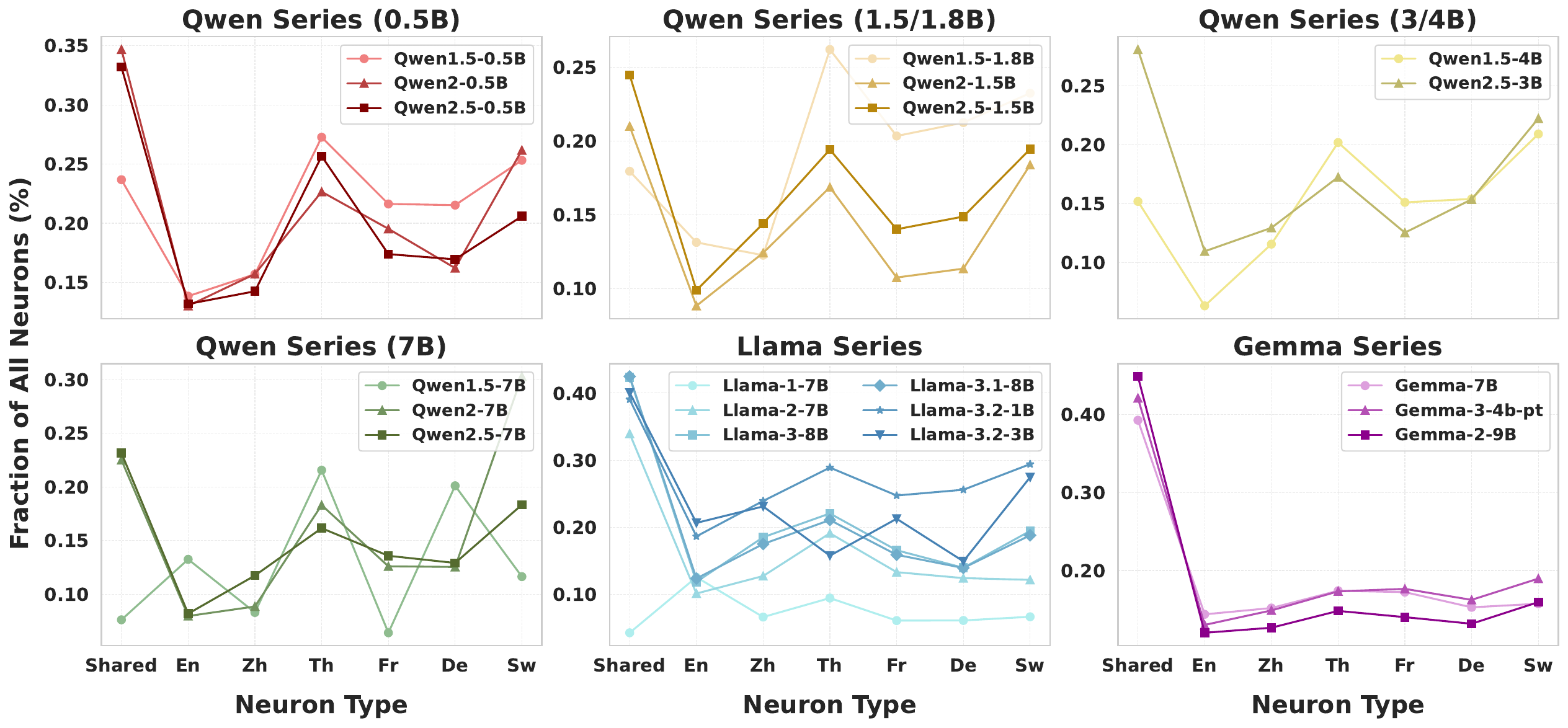}
    \vspace{-10pt}
    \caption{
    Neuron distribution across language-shared and language-exclusive neurons for six languages (En, Zh, Th, Fr, Dr, Sw) in various model series and scales. 
    For each model, we present the fraction of shared neurons and exclusive neurons to the total number of neurons. 
    }
    \label{fig:neuron_distribution}
    \vspace{-10pt}
\end{figure}
\vspace{-5pt}

\paragraph{Language-related neurons account for only a small proportion in LLMs.} To develop a preliminary understanding of language-shared and language-exclusive neurons, we begin by analyzing the distribution of shared and language-exclusive parameters across all neurons within the model. For each language, we compute the proportion of language-shared and language-exclusive neurons relative to the total number of neurons in the model. Specifically, we calculate the ratios $\mathcal{N}_{\mathrm{shared}} / |\mathcal{LLM}|$ and $\mathcal{N}^{\ell}_{\mathrm{exclusive}} / |\mathcal{LLM}|$, where $\ell$ denotes a specific language. The results, illustrated in Figure~\ref{fig:neuron_distribution}, encompass six languages across multiple model series.
It shows that only a small fraction of neurons, often fewer than $1\%$, play a critical role in processing language, underscoring the sparsity and selectivity of language-relevant neural activations. Furthermore, the quantities of language-shared and language-exclusive neurons are of similar magnitude, each coarsely estimated at around $0.3\%$ of the total number of neurons in the LLM.

\begin{figure}[t]
    \centering
    \vspace{-5pt}
    \includegraphics[width=1\textwidth]{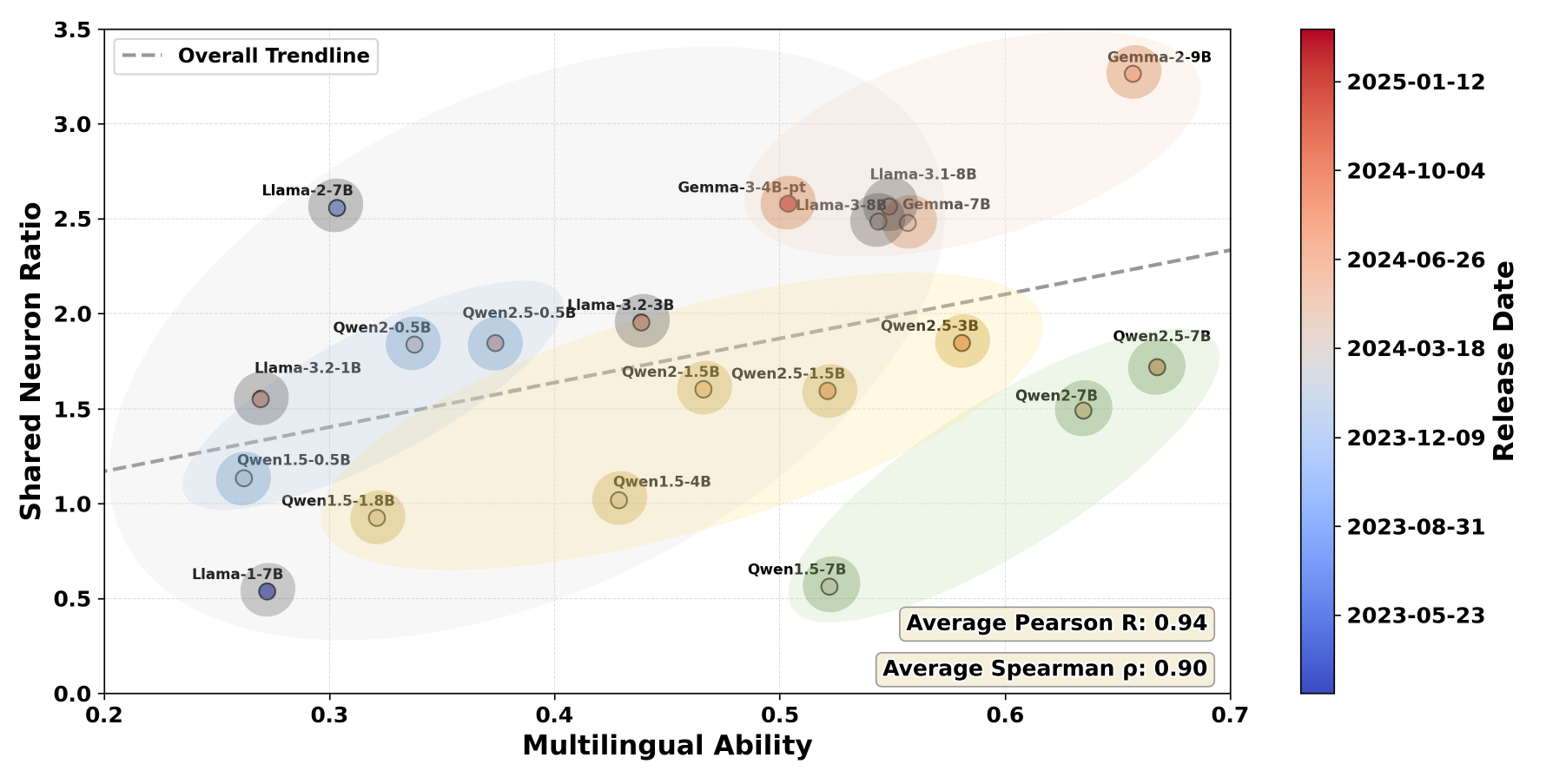}
    \vspace{-15pt}
    \caption{The relationship between multilingual ability and shared neuron ratio (as defined in Equation \ref{equ:share}) across various models. Each point represents a model, color-coded by its release date. Shaded regions indicate groups of models within the same series and of comparable scale. The gray dashed line (\textcolor{gray}{- - - -}) illustrates the overall trend: as models evolve, those with greater multilingual capabilities tend to exhibit a higher proportion of shared neurons.}
    \label{fig:shared_neuron_percentage}
    \vspace{-10pt}
\end{figure}

To further explore the evolution of shared and exclusive neurons across models and over time, we compute the overall shared neuron ratio for each model, as defined in Equation~\ref{equ:share}, relate it to multilingual performance measured by MMMLU and MGSM, and present the results in Figure~\ref{fig:shared_neuron_percentage}.

\vspace{-5pt}

\paragraph{The proportion of shared neurons increases with model evolution.}

We first group models from the same series and with similar parameter scales, as indicated by the shaded color regions in Figure \ref{fig:shared_neuron_percentage}. 
Within each group (e.g., Qwen1.5-7B, Qwen2-7B, and Qwen2.5-7B), we observe a steady and consistent increase in the shared-to-exclusive neuron ratio across generations. 
This growth closely parallels improvements in the model's multilingual ability, with an average Pearson correlation coefficient of $R = 0.92$ and a Spearman rank correlation of $\rho = 0.88$, indicating a strong and reliable relationship. 
In other words, later generations within the same series show a strong trend toward engaging more shared neurons for processing different languages.

\paragraph{The increase of shared neuron proportion generalizes across model families.}
Beyond individual model series, we observe that the positive relationship between the proportion of shared neurons and multilingual capability generally persists across different model families, as illustrated by the gray dashed line (\textcolor{gray}{- - - -}) in Figure \ref{fig:shared_neuron_percentage}.
Despite differences in architecture design and pretraining corpora, models with stronger multilingual ability tend to activate a larger proportion of shared neurons. 
For instance, the Gemma series exhibits both the most strong multilingual performance and the highest shared-to-exclusive neuron ratio. 
This consistency across diverse architectures suggests that progressively leveraging shared neurons may be a general strategy adopted by multilingual LLMs, regardless of their specific design choices.

\subsection{Analysis on Language Agnostic Score}

\begin{wrapfigure}{r}{0.66\textwidth}
    \vspace{-0.2cm}
    \centering
    \vspace{-0.2cm}
    \includegraphics[width=0.65\textwidth]{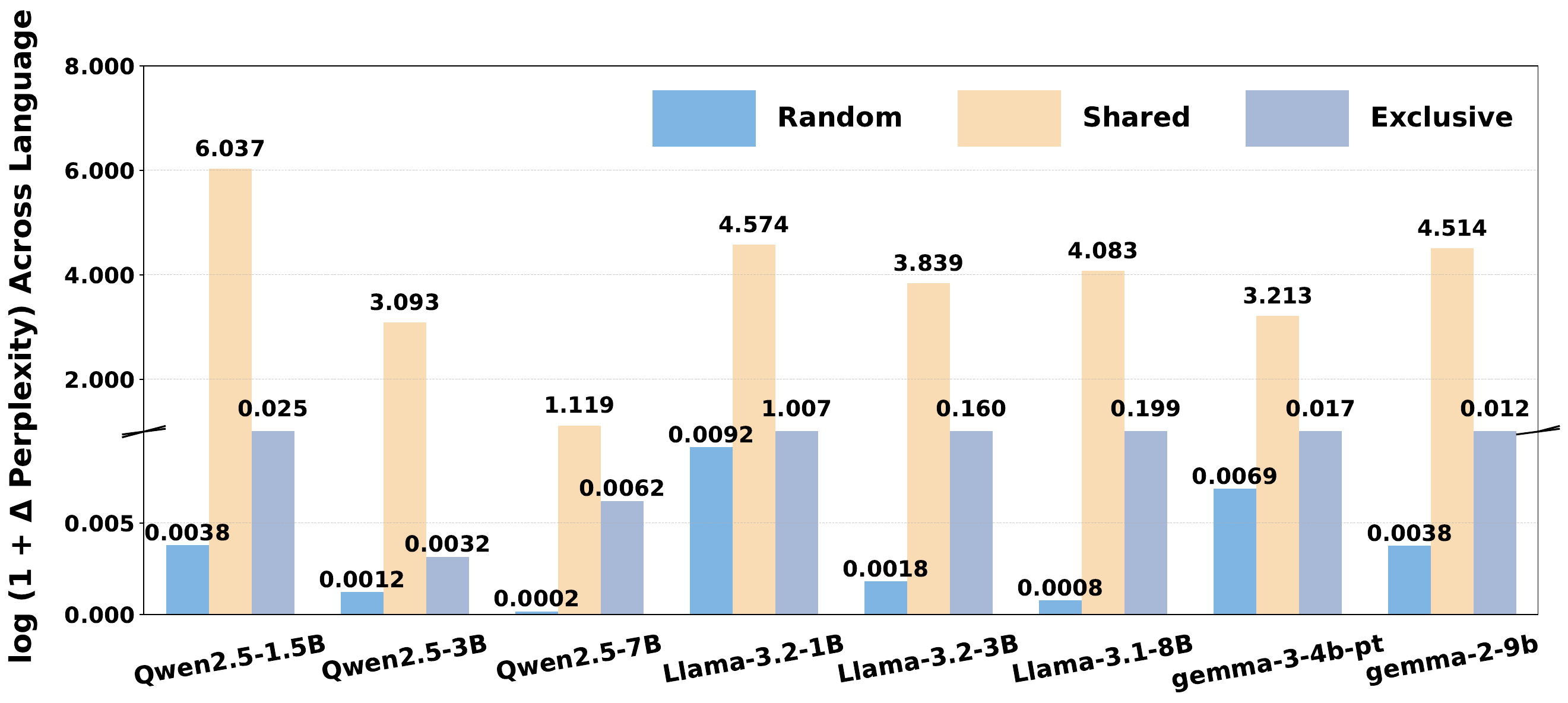}
    \caption{
    Perplexity changes caused by deactivating random neuron sets (Random), language-shared neurons (Shared) and language-exclusive neurons (Exclusive). Notice that Random deactivation barely affects models' perplexity, while Shared and Exclusive deactivation break the models' abilities.
    }
    \label{fig:random_deactivation}
    \vspace{-0.2cm}
\end{wrapfigure}

The above observations raise a further question: whether the shared neurons not only occupy a larger proportion of the language-related neuron set, but also \textit{contribute more critically to multilingual processing}, effectively functioning as language-agnostic neurons. To address this question, we investigate how the language-shared neurons importance, i.e., language agnostic score defined in Equation \ref{equ:agnostic}, evolves alongside multilingual capability across different generations of large language models.

\begin{figure}[t]
    \centering
    \vspace{-5pt}
    \includegraphics[width=1\textwidth]{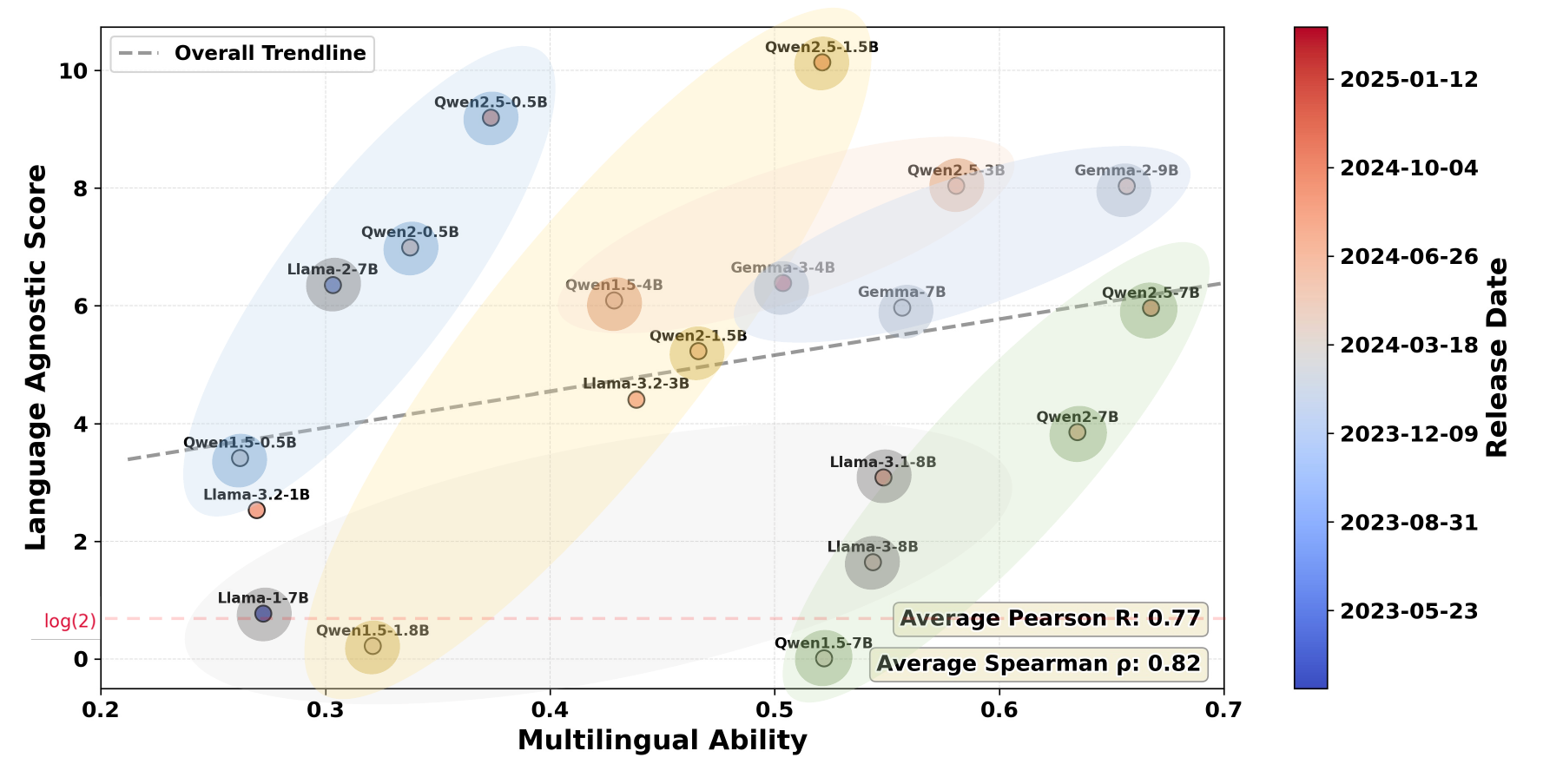}
    \vspace{-15pt}
    \caption{
The relationship between multilingual ability and language-agnostic score (as defined in Equation \ref{equ:agnostic}) across various language models.
Each point represents a model, colored by model size. The red dashed line (\textcolor{red}{- - - -}) indicates where shared neuron influence surpasses that of exclusive neurons, i.e., $\mathrm{Imp}=1$ and $\mathrm{Language\;Agnostic\;Score} = \log (2)$. Shaded regions group models within the same series and of similar scale, while the dashed line (\textcolor{gray}{- - - -}) indicates the overall trend: as LLMs evolve, successive generations with enhanced multilingual capabilities tend to achieve higher language-agnostic scores. This trend suggests that shared neurons increasingly support not only multilingual processing but also the emergence of more language-agnostic, abstract thought.
    }
    \label{fig:deactivation}
    \vspace{-5pt}
\end{figure}

\paragraph{Deactivating shared and exclusive neurons both leads to model degradation.}

Before contrasting language-shared neurons with language-exclusive neurons, we conduct a control experiment in which we deactivate an equal number of randomly selected neurons, matching the quantity of both language-shared and language-exclusive neurons. As illustrated in Figure~\ref{fig:random_deactivation}, this random deactivation results in minimal changes in perplexity across languages. In contrast, deactivating language-shared or language-exclusive neurons leads to significant performance degradation. These results confirm that the identified language-related neurons are indeed specialized for language processing, and that our neuron importance metrics are robust to random perturbations.

To further investigate whether language-shared neurons evolve into language-agnostic neurons, we analyze the evolution of the language-agnostic score, as defined in Equation~\ref{equ:agnostic}, in relation to the models' multilingual capabilities, as shown in Figure~\ref{fig:deactivation}.

\vspace{-2mm}
\paragraph{Shared neurons in early-stage models reflect superficial overlap without supporting higher-level cognition.}
In earlier models such as the Qwen1.5 series and LLaMA-1-7B, deactivating shared neurons has a comparable effect to deactivating exclusive neurons, with language-agnostic scores around 1.
This suggests that in early-stage models, shared neurons have the similar importance with exclusive neuron, and shared neurons largely reflect superficial overlaps between language-related neuron across languages, rather than representing a distinct, functionally meaningful shared space.

\vspace{-2mm}
\paragraph{Shared neurons in recent models become central and exhibit language-agnostic properties.}
In contrast, recent models, such as those in the Qwen2.5 series, exhibit a dramatically different pattern. Deactivating shared neurons leads to a sharp and disproportionate increase in perplexity across all languages, often several orders of magnitude greater than the increase caused by removing language-exclusive neurons.
In other words, shared neurons contribute far more critically to multilingual processing than exclusive neurons, despite both being part of the language-related neuron set. 
This disproportionate degradation reveals that shared neurons in recent models have evolved beyond serving merely as intersections of language-specific components; they now fulfill more fundamental, language-agnostic roles.
If such shared neurons have indeed evolved into language-agnostic neurons, they may be operating within a conceptual space that abstracts away from surface-level linguistic variations. 
Such a space would allow the model to perform high-level reasoning, semantic alignment, and cross-lingual generalization—hallmarks of abstract thought in multilingual LLMs.

\section{Multilingual Enhancement via Neuron-Targeted Training}

\begin{table}[t]
\centering
\renewcommand{\arraystretch}{1.3}
\caption{Multilingual performance improvements on MGSM (primarily involving abstract thought) and MMMLU (requiring both abstract thought and domain knowledge) across five languages. Models were trained only on 100,000 general documents without reasoning-related data and evaluated using Llama-3.1-8B (high language-agnostic), Llama-3.2-3B (medium language-agnostic), and Llama-3.2-1B (low language-agnostic) under various targeted neuron tuning strategies.}
\footnotesize
\setlength{\tabcolsep}{1.2pt}
\scalebox{0.9}{
\begin{tabular}{c|l|cccccc|cccccc}
\toprule
 \qquad & \multirow{2}{*}{\normalsize{\textbf{Neuron}}} & \multicolumn{6}{c|}{\normalsize{\textbf{MGSM}}} & \multicolumn{6}{c}{\normalsize{\textbf{MMMLU}}} \\
& & {Zh} & {Fr} & {De} & {Th} & {Sw} & $\boldsymbol{\Delta}_{Avg}$ & {Zh} & {Fr} & {De} & {Th} & {Sw} & $\boldsymbol{\Delta}_{Avg}$ \\
\midrule
\rowcolor{lightgray}\cellcolor{white}\multirow{4}{*}{\rotatebox{90}{\textbf{Llama-3.1-8B}}} 
& None                  & 52.4 & 51.6 & 54.4 & 46.8 & 38.8 & -   & 53.8 & 58.4 & 56.9 & 48.8 & 40.9 & - \\
& Shared              &   52.0$^{\color{-}-0.4}$   &  52.8$^{\color{+}+1.2}$    &  55.6$^{\color{+}+1.2}$   &  45.6$^{\color{-}-1.2}$    &   39.6$^{\color{+}+0.8}$   & 0.3        &   54.6$^{\color{+}+0.8}$     &   57.2$^{\color{-}-1.2}$    &   56.5$^{\color{-}-0.4}$    &    48.9$^{\color{+}+0.1}$    & 42.3$^{\color{+}+1.4}$  & 0.1 \\
& \cellcolor{lightblue}Exclusive             &  \cellcolor{lightblue}56.8$^{\color{+}+4.4}$    &   \cellcolor{lightblue}57.2$^{\color{+}+5.6}$   &  \cellcolor{lightblue}57.2$^{\color{+}+2.8}$    &   \cellcolor{lightblue}50.4$^{\color{+}+3.6}$   &  \cellcolor{lightblue}42.4$^{\color{+}+3.6}$    & \cellcolor{lightblue}\textbf{4.0}    &    \cellcolor{lightblue}55.6$^{\color{+}+1.8}$    &   \cellcolor{lightblue}59.2$^{\color{+}+0.8}$     &   \cellcolor{lightblue}59.1$^{\color{+}+2.2}$     &   \cellcolor{lightblue}49.9$^{\color{+}+1.1}$     &   \cellcolor{lightblue}43.7$^{\color{+}+2.8}$     &  \cellcolor{lightblue}\textbf{1.7}  \\
& Random                &   50.4$^{\color{-}-2.0}$    &   51.2$^{\color{-}-0.4}$   &  54.4 $^{\color{-}-0.0}$   &  47.2$^{\color{+}+0.4}$    &  37.6$^{\color{-}-1.2}$    &  -0.6   &    52.4$^{\color{-}-1.4}$   &   58.3$^{\color{-}-0.1}$    &   57.2$^{\color{+}+0.3}$     &  47.1$^{\color{-}-1.7}$     &   41.3$^{\color{+}+0.4}$     &  -0.5 \\

\midrule

\rowcolor{lightgray}\cellcolor{white}\multirow{4}{*}{\rotatebox{90}{\textbf{Llama-3.2-3B}}} 
& None                  & 40.8 & 42.4 & 57.2 & 35.2 & 30.8 & -   & 45.2 & 49.0 & 47.1 & 40.6 & 34.1 & - \\
& \cellcolor{lightblue}Shared              & \cellcolor{lightblue}42.8$^{\color{+}+2.0}$ & \cellcolor{lightblue}45.6$^{\color{+}+3.2}$ & \cellcolor{lightblue}66.4$^{\color{+}+9.2}$ & \cellcolor{lightblue}40.4$^{\color{+}+5.2}$ & \cellcolor{lightblue}39.6$^{\color{+}+8.8}$ & \cellcolor{lightblue}\textbf{5.7} & \cellcolor{lightblue}44.9$^{\color{-}-0.3}$ & \cellcolor{lightblue}49.8$^{\color{+}+0.8}$ & \cellcolor{lightblue}47.3$^{\color{+}+0.2}$ & \cellcolor{lightblue}41.0$^{\color{+}+0.4}$ & \cellcolor{lightblue}34.8$^{\color{+}+0.7}$ & \cellcolor{lightblue}\textbf{0.4} \\
& Exclusive             & 42.4$^{\color{+}+1.6}$ & 43.2$^{\color{+}+0.8}$ & 65.6$^{\color{+}+8.4}$ & 37.2$^{\color{+}+2.0}$ & 36.0$^{\color{+}+5.2}$ & 3.6 & 44.9$^{\color{-}-0.3}$ & 48.9$^{\color{-}-0.1}$ & 47.1$^{\color{+}+0.0}$ & 40.9$^{\color{+}+0.3}$ & 34.7$^{\color{+}+0.6}$ & 0.1 \\
& Random                & 40.4$^{\color{-}-0.4}$ & 41.6$^{\color{-}-0.8}$ & 63.2$^{\color{+}+6.0}$ & 34.8$^{\color{-}-0.4}$ & 30.0$^{\color{-}-0.8}$ & 0.7 & 44.5$^{\color{-}-0.7}$ & 49.0$^{\color{+}+0.0}$ & 46.9$^{\color{-}-0.2}$ & 40.3$^{\color{-}-0.3}$ & 34.1$^{\color{+}+0.0}$ & -0.2 \\

\midrule

\rowcolor{lightgray}\cellcolor{white}\multirow{4}{*}{\rotatebox{90}{\textbf{Llama-3.2-1B}}}
& None                  & 26.4 & 26.0 & 29.2 & 20.0 & 22.8 & -   & 29.0 & 27.8 & 28.8 & 28.8 & 26.6 & - \\
& \cellcolor{lightblue}Shared & \cellcolor{lightblue}30.0$^{\color{+}+3.6}$ &\cellcolor{lightblue} 30.4$^{\color{+}+4.4}$ & \cellcolor{lightblue}30.8$^{\color{+}+1.6}$ & \cellcolor{lightblue}22.4$^{\color{+}+2.4}$ & \cellcolor{lightblue}26.4$^{\color{+}+3.6}$ & \cellcolor{lightblue}3.1 & \cellcolor{lightblue}29.2$^{\color{+}+0.2}$ & \cellcolor{lightblue}28.7$^{\color{+}+0.9}$ & \cellcolor{lightblue}29.5$^{\color{+}+0.7}$ & \cellcolor{lightblue}29.4$^{\color{+}+0.6}$ & \cellcolor{lightblue}26.8$^{\color{+}+0.2}$ & \cellcolor{lightblue}\textbf{0.5} \\
& \cellcolor{lightblue}Exclusive             & \cellcolor{lightblue}27.6$^{\color{+}+1.2}$ & \cellcolor{lightblue}30.0$^{\color{+}+4.0}$ & \cellcolor{lightblue}34.4$^{\color{+}+5.2}$ & \cellcolor{lightblue}23.2$^{\color{+}+3.2}$ & \cellcolor{lightblue}30.4$^{\color{+}+7.6}$ & \cellcolor{lightblue}\textbf{4.2} & \cellcolor{lightblue}29.0$^{\color{-}-0.0}$ & \cellcolor{lightblue}28.0$^{\color{+}+0.2}$ & \cellcolor{lightblue}29.3$^{\color{+}+0.5}$ & \cellcolor{lightblue}28.2$^{\color{-}-0.6}$ & \cellcolor{lightblue}26.8$^{\color{+}+0.2}$ & \cellcolor{lightblue}0.1 \\
& Random                & 26.8$^{\color{+}+0.4}$ & 26.4$^{\color{-}-0.4}$ & 29.6$^{\color{+}+0.4}$ & 21.2$^{\color{+}+1.2}$ & 26.4$^{\color{+}+3.6}$ & 1.0 & 28.8$^{\color{-}-0.2}$ & 28.3$^{\color{+}+0.5}$ & 29.1$^{\color{+}+0.3}$ & 28.6$^{\color{-}-0.2}$ & 26.8$^{\color{+}+0.2}$ & 0.1 \\

\bottomrule
\end{tabular}}
\label{tab:tuning_neuron_gsm8k_mmlu}
\end{table}

\subsection{Language Agnostic Score Guided Multilingual Enhance}

Inspired by above insights, we propose various targeted neuron training methods to enhance models' multilingual capability according to their language agnostic score.
\vspace{-3mm}
\paragraph{LLMs with low language agnostic score can train any language-related neurons.} 

These models exhibit limited multilingual capabilities, indicating that all language-related neurons require improvement. To enhance their performance across languages, we propose training all language-related neurons, whether they are shared across languages or specific to individual ones.
\vspace{-3mm}
\paragraph{LLMs with middle language agnostic score should train language-shared neurons.}

These models demonstrate a degree of multilingual capability; however, the language-shared neurons have not yet evolved to become truly language-agnostic. Given that language-shared neurons are more prevalent than language-exclusive ones in these models, it is essential to further train and refine them to more effectively enhance the models' multilingual performance.

\vspace{-3mm}
\paragraph{LLMs with high language agnostic score should train language-exclusive neurons} The language-shared neurons in these models have evolved into language-agnostic neurons, responsible for abstract thought. They are already well-trained and offer limited room for further improvement. Therefore, to enhance multilingual performance, it is necessary to focus on training the language-exclusive neurons in these LLMs.

\subsection{Experiment Setup}
\label{subsec:training_setup}
\vspace{-2mm}
\paragraph{Dataset}

To further validate our hypothesis and explore how to utilize our findings to efficiently enhance multilingual capability in LLMs, we conduct continuous pretraining on specific neurons using multilingual corpora. 
Specifically, we construct a training set by sampling 100,000 examples per language from a mixture of three widely used multilingual datasets: Culturax \citep{CulturaX}, MADLAD \citep{MADLAD}, and Wikipedia \citep{Wiki}. 
\vspace{-2mm}
\paragraph{Training Settings}

We utilize Llama3.2-1B~\citep{grattafiori2024llama}, Llama3.2-3B, and Lamma-3.1-8B as representative LLMs with low, medium, and high language-agnostic scores, respectively. We conduct experiments under three training settings: language-shared neurons, language-exclusive neurons, and an equal number of randomly selected neurons. To evaluate multilingual capability, we employ the MMMLU and MGSM benchmarks.

\vspace{-2mm}
\paragraph{Experiment Results}

Table \ref{tab:tuning_neuron_gsm8k_mmlu} demonstrates that the multilingual capabilities of language models can be effectively enhanced through targeted neuron-specific tuning. For Llama-3.2-1B, which exhibits a relatively low language-agnostic score, tuning both shared and exclusive neurons significantly improves the model’s cross-lingual reasoning performance, yielding average gains of 3.1 and 4.2 points on the MGSM benchmark, respectively. In the case of Llama-3.2-3B, which has a moderate language-agnostic score, tuning language-shared neurons results in the greatest performance improvement—an average gain of 5.7 points on MGSM. This is likely because these neurons are more numerous and less well-trained than exclusive ones. Finally, for Llama-3.2-8B, which already possesses a high language-agnostic score, the language-shared neurons appear to be sufficiently trained; thus, tuning exclusive neurons leads to further enhancement of multilingual performance, with an observed improvement of 4.0 points on MGSM.
Compared to the improvement observed on MGSM, the performance gain on MMLU—which relies more heavily on knowledge extraction—is relatively smaller. This suggests that our training approach primarily enhances the model's thinking capabilities rather than its factual recall. Moreover, since we exclusively utilize general documents without incorporating reasoning-specific data, the substantial improvement further validates the effectiveness of our neuron-targeted training methodology.


\section{Related Work}

\paragraph{Thinking Language of LLMs}

Large language models (LLMs) \citep{touvron2023llama2, openai2023gpt4, yang2024qwen2technicalreport, touvron2023llama, team2025gemma, team2024gemma2} demonstrate strong multilingual reasoning and transfer abilities \citep{pires-etal-2019-multilingual, wu-dredze-2019-beto, hardalov2021fewshotcrosslingualstancedetection, nooralahzadeh-etal-2020-zero}, raising questions about whether these models operate in a language-agnostic or language-specific concept space \citep{nanda2023progressmeasuresgrokkingmechanistic, schut2025multilingual, how_do_handle}, and which language would the model ``think'' in. 
One stream of work supports the hypothesis that LLMs ``think'' in a concept space centered on the predominant training language. \citet{zhong2024englishcentricllmslanguagemultilingual} analyzed LLMs trained predominantly on English or Japanese \citep{fujii2024continualpretrainingcrosslingualllm, llmjp2024llmjpcrossorganizationalprojectresearch} for their mainly activated languages; \citet{fierro2025multilinguallanguagemodelsremember} showed language dependence in object retrieval; and \citet{schut2025multilingualllmsthinkenglish}, found representations align more closely with English even on foreign inputs.
On the other hand, a language-agnostic view is supported by either probing studies  \citep{pires-etal-2019-multilingual, stanczak-etal-2022-neurons}, neuron-level manipulations \citep{dumas2024how, brinkmann2025largelanguagemodelsshare} or both \citep{wu2025semantichubhypothesislanguage, wendler2024llamas}.
Our work falls in line with \citet{dumas2024how, wendler2024llamas, wu2025semantichubhypothesislanguage}, with more fine-grained neuron-level results and a novel activation-and-training-based analysis method.
\vspace{-2mm}
\paragraph{Multilingual Enhancement}

Early-on, multilingual enhancement is mainly approached from pre-training in works such as XLM, XLM-R\citep{conneau-etal-2020-unsupervised, lample2019crosslinguallanguagemodelpretraining} and M-BERT \citep{DBLP:journals/corr/abs-1810-04805}. More post-training work, ranging from continual pre-training \citep{zhang2021cpm2largescalecosteffectivepretrained, cui2024efficienteffectivetextencoding, chen2023lifelonglanguagepretrainingdistributionspecialized, husain2024romansetuefficientlyunlockingmultilingual, kuulmets2024teachingllamanewlanguage} to fine-tuning \citep{muennighoff2023crosslingualgeneralizationmultitaskfinetuning, chen2023lifelonglanguagepretrainingdistributionspecialized, ahuja2024sphinxsampleefficientmultilingual, lai2023okapiinstructiontunedlargelanguage, indurthi2024improvingmultilingualinstructionfinetuning, lai2024mcotmultilingualinstructiontuning, zhao2024adamergexcrosslingualtransferlarge} have emerged to effectively improve models' multilingual abilities, though rather sensitive to training corpus and settings. 
A parallel body of work focuses on prompt-based methods, either leaning on language alignment \citep{zhang2024autocapautomaticcrosslingualalignment, etxaniz2023multilinguallanguagemodelsthink} or instruction-following and attention \citep{wang2025largelanguagemodelsgood,huang2023languagescreatedequalllms}. However, \citet{liu2024translation} points out the suboptimality in translation-based prompting pipelines.
Our neuron-specific tuning strategy answers the academic call \citep{liu2024translation, liu2025translationneedstudysolving} for a more comprehensive approach to multilingual enhancement than translation-based prompting, and provides a more efficient and task-neutral alternative than the post-training based methods. 

\section{Conclusion \& Discussion}
\label{sec:conclusion}

In this work, we explore the emergence of abstract thought in large language models through the lens of neuron behavior. By identifying and categorizing language-related neurons as either shared or exclusive, we uncover a consistent trend across model development: shared neurons not only increase in proportion but also grow in functional importance, eventually forming a compact yet critical set of language-agnostic neurons. These neurons underpin the model’s ability to generalize across languages and support abstract reasoning that transcends linguistic boundaries.
Motivated by this insight, we introduce neuron-specific training strategies that adapt to the developmental stage of an LLM, whether or not it exhibits language-agnostic behavior. Extensive experiments confirm that our targeted training approach effectively enhances multilingual performance across diverse models. We believe this neuron-centric perspective opens new avenues for understanding and improving the generalization capabilities of LLMs in multilingual and cross-lingual contexts.

Our study is limited by computational resources, which restricts evaluation on the larger LLMs and prevents full exploration of the potential of neuron-centric training at larger scales. We leave these directions for future work. Nonetheless, our findings shed light on the emergence of multilingual and abstract reasoning in LLMs, which may promote language equity but also raise risks like cross-lingual misinformation, calling for responsible deployment.


\bibliographystyle{acl}
\bibliography{reference}

\begin{thebibliography}{66}
\expandafter\ifx\csname natexlab\endcsname\relax\def\natexlab#1{#1}\fi

\bibitem[{Abadji et~al.(2022)Abadji, Su{\'{a}}rez, Romary, and Sagot}]{oscar}
Julien Abadji, Pedro Javier~Ortiz Su{\'{a}}rez, Laurent Romary, and
  Beno{\^{\i}}t Sagot. 2022.
\newblock Towards a cleaner document-oriented multilingual crawled corpus.
\newblock In \emph{{LREC}}, pages 4344--4355. European Language Resources
  Association.

\bibitem[{Ahuja et~al.(2024)Ahuja, Tanmay, Chauhan, Patra, Aggarwal, Corro,
  Mitra, Dhamecha, Awadallah, Choudhary, Chaudhary, and
  Sitaram}]{ahuja2024sphinxsampleefficientmultilingual}
Sanchit Ahuja, Kumar Tanmay, Hardik~Hansrajbhai Chauhan, Barun Patra, Kriti
  Aggarwal, Luciano~Del Corro, Arindam Mitra, Tejas~Indulal Dhamecha, Ahmed
  Awadallah, Monojit Choudhary, Vishrav Chaudhary, and Sunayana Sitaram. 2024.
\newblock \href {http://arxiv.org/abs/2407.09879} {sphinx: Sample efficient
  multilingual instruction fine-tuning through n-shot guided prompting}.

\bibitem[{Bai et~al.(2023)Bai, Bai, Chu, Cui, Dang, Deng, Fan, Ge, Han, Huang
  et~al.}]{bai2023qwen}
Jinze Bai, Shuai Bai, Yunfei Chu, Zeyu Cui, Kai Dang, Xiaodong Deng, Yang Fan,
  Wenbin Ge, Yu~Han, Fei Huang, et~al. 2023.
\newblock Qwen technical report.
\newblock \emph{arXiv preprint arXiv:2309.16609}.

\bibitem[{Brinkmann et~al.(2025)Brinkmann, Wendler, Bartelt, and
  Mueller}]{brinkmann2025largelanguagemodelsshare}
Jannik Brinkmann, Chris Wendler, Christian Bartelt, and Aaron Mueller. 2025.
\newblock \href {http://arxiv.org/abs/2501.06346} {Large language models share
  representations of latent grammatical concepts across typologically diverse
  languages}.

\bibitem[{Chen et~al.(2023)Chen, Zhou, Du, Huang, Laudon, Chen, and
  Cu}]{chen2023lifelonglanguagepretrainingdistributionspecialized}
Wuyang Chen, Yanqi Zhou, Nan Du, Yanping Huang, James Laudon, Zhifeng Chen, and
  Claire Cu. 2023.
\newblock \href {http://arxiv.org/abs/2305.12281} {Lifelong language
  pretraining with distribution-specialized experts}.

\bibitem[{Cobbe et~al.(2021)Cobbe, Kosaraju, Bavarian, Chen, Jun, Kaiser,
  Plappert, Tworek, Hilton, Nakano, Hesse, and Schulman}]{cobbe2021gsm8k}
Karl Cobbe, Vineet Kosaraju, Mohammad Bavarian, Mark Chen, Heewoo Jun, Lukasz
  Kaiser, Matthias Plappert, Jerry Tworek, Jacob Hilton, Reiichiro Nakano,
  Christopher Hesse, and John Schulman. 2021.
\newblock Training verifiers to solve math word problems.
\newblock \emph{arXiv preprint arXiv:2110.14168}.

\bibitem[{Conneau et~al.(2020)Conneau, Khandelwal, Goyal, Chaudhary, Wenzek,
  Guzm{\'a}n, Grave, Ott, Zettlemoyer, and
  Stoyanov}]{conneau-etal-2020-unsupervised}
Alexis Conneau, Kartikay Khandelwal, Naman Goyal, Vishrav Chaudhary, Guillaume
  Wenzek, Francisco Guzm{\'a}n, Edouard Grave, Myle Ott, Luke Zettlemoyer, and
  Veselin Stoyanov. 2020.
\newblock \href {https://doi.org/10.18653/v1/2020.acl-main.747} {Unsupervised
  cross-lingual representation learning at scale}.
\newblock In \emph{Proceedings of the 58th Annual Meeting of the Association
  for Computational Linguistics}, pages 8440--8451, Online. Association for
  Computational Linguistics.

\bibitem[{Cui et~al.(2024)Cui, Yang, and
  Yao}]{cui2024efficienteffectivetextencoding}
Yiming Cui, Ziqing Yang, and Xin Yao. 2024.
\newblock \href {http://arxiv.org/abs/2304.08177} {Efficient and effective text
  encoding for chinese llama and alpaca}.

\bibitem[{Devlin et~al.(2018)Devlin, Chang, Lee, and
  Toutanova}]{DBLP:journals/corr/abs-1810-04805}
Jacob Devlin, Ming{-}Wei Chang, Kenton Lee, and Kristina Toutanova. 2018.
\newblock \href {http://arxiv.org/abs/1810.04805} {{BERT:} pre-training of deep
  bidirectional transformers for language understanding}.
\newblock \emph{CoRR}, abs/1810.04805.

\bibitem[{Dumas et~al.(2024)Dumas, Veselovsky, Monea, West, and
  Wendler}]{dumas2024how}
Cl{\'e}ment Dumas, Veniamin Veselovsky, Giovanni Monea, Robert West, and Chris
  Wendler. 2024.
\newblock \href {https://openreview.net/forum?id=0ku2hIm4BS} {How do llamas
  process multilingual text? a latent exploration through activation patching}.
\newblock In \emph{ICML 2024 Workshop on Mechanistic Interpretability}.

\bibitem[{Etxaniz et~al.(2023)Etxaniz, Azkune, Soroa, de~Lacalle, and
  Artetxe}]{etxaniz2023multilinguallanguagemodelsthink}
Julen Etxaniz, Gorka Azkune, Aitor Soroa, Oier~Lopez de~Lacalle, and Mikel
  Artetxe. 2023.
\newblock \href {http://arxiv.org/abs/2308.01223} {Do multilingual language
  models think better in english?}

\bibitem[{Fierro et~al.(2025)Fierro, Foroutan, Elliott, and
  Søgaard}]{fierro2025multilinguallanguagemodelsremember}
Constanza Fierro, Negar Foroutan, Desmond Elliott, and Anders Søgaard. 2025.
\newblock \href {http://arxiv.org/abs/2410.14387} {How do multilingual language
  models remember facts?}

\bibitem[{Frankle and Carbin(2018)}]{frankle2018lottery}
Jonathan Frankle and Michael Carbin. 2018.
\newblock The lottery ticket hypothesis: Finding sparse, trainable neural
  networks.
\newblock In \emph{International Conference on Learning Representations}.

\bibitem[{Fujii et~al.(2024)Fujii, Nakamura, Loem, Iida, Ohi, Hattori, Shota,
  Mizuki, Yokota, and Okazaki}]{fujii2024continualpretrainingcrosslingualllm}
Kazuki Fujii, Taishi Nakamura, Mengsay Loem, Hiroki Iida, Masanari Ohi, Kakeru
  Hattori, Hirai Shota, Sakae Mizuki, Rio Yokota, and Naoaki Okazaki. 2024.
\newblock \href {http://arxiv.org/abs/2404.17790} {Continual pre-training for
  cross-lingual llm adaptation: Enhancing japanese language capabilities}.

\bibitem[{Gemma~Team et~al.(2025)Gemma~Team, Kamath, Ferret, Pathak, Vieillard,
  Merhej, Perrin, Matejovicova, Ram{\'e}, Rivi{\`e}re et~al.}]{team2025gemma}
Gemma Gemma~Team, Aishwarya Kamath, Johan Ferret, Shreya Pathak, Nino
  Vieillard, Ramona Merhej, Sarah Perrin, Tatiana Matejovicova, Alexandre
  Ram{\'e}, Morgane Rivi{\`e}re, et~al. 2025.
\newblock Gemma 3 technical report.
\newblock \emph{arXiv preprint arXiv:2503.19786}.

\bibitem[{Gemma~Team et~al.(2024{\natexlab{a}})Gemma~Team, Mesnard, Hardin,
  Dadashi, Bhupatiraju, Pathak, Sifre, Rivi{\`e}re, Kale, Love
  et~al.}]{team2024gemma}
Gemma Gemma~Team, Thomas Mesnard, Cassidy Hardin, Robert Dadashi, Surya
  Bhupatiraju, Shreya Pathak, Laurent Sifre, Morgane Rivi{\`e}re, Mihir~Sanjay
  Kale, Juliette Love, et~al. 2024{\natexlab{a}}.
\newblock Gemma: Open models based on gemini research and technology.
\newblock \emph{arXiv preprint arXiv:2403.08295}.

\bibitem[{Gemma~Team et~al.(2024{\natexlab{b}})Gemma~Team, Riviere, Pathak,
  Sessa, Hardin, Bhupatiraju, Hussenot, Mesnard, Shahriari, Ram{\'e}
  et~al.}]{team2024gemma2}
Gemma Gemma~Team, Morgane Riviere, Shreya Pathak, Pier~Giuseppe Sessa, Cassidy
  Hardin, Surya Bhupatiraju, L{\'e}onard Hussenot, Thomas Mesnard, Bobak
  Shahriari, Alexandre Ram{\'e}, et~al. 2024{\natexlab{b}}.
\newblock Gemma 2: Improving open language models at a practical size.
\newblock \emph{arXiv preprint arXiv:2408.00118}.

\bibitem[{Grattafiori et~al.(2024)Grattafiori, Dubey, Jauhri, Pandey, Kadian,
  Al-Dahle, Letman, Mathur, Schelten, Vaughan et~al.}]{grattafiori2024llama}
Aaron Grattafiori, Abhimanyu Dubey, Abhinav Jauhri, Abhinav Pandey, Abhishek
  Kadian, Ahmad Al-Dahle, Aiesha Letman, Akhil Mathur, Alan Schelten, Alex
  Vaughan, et~al. 2024.
\newblock The llama 3 herd of models.
\newblock \emph{arXiv preprint arXiv:2407.21783}.

\bibitem[{Guo et~al.(2020)Guo, Dai, Vrandecic, and Al{-}Rfou}]{Wiki}
Mandy Guo, Zihang Dai, Denny Vrandecic, and Rami Al{-}Rfou. 2020.
\newblock Wiki-40b: Multilingual language model dataset.
\newblock In \emph{{LREC}}, pages 2440--2452. European Language Resources
  Association.

\bibitem[{Hardalov et~al.(2021)Hardalov, Arora, Nakov, and
  Augenstein}]{hardalov2021fewshotcrosslingualstancedetection}
Momchil Hardalov, Arnav Arora, Preslav Nakov, and Isabelle Augenstein. 2021.
\newblock \href {http://arxiv.org/abs/2109.06050} {Few-shot cross-lingual
  stance detection with sentiment-based pre-training}.

\bibitem[{Hendrycks et~al.(2021)Hendrycks, Burns, Basart, Zou, Mazeika, Song,
  and Steinhardt}]{hendrycks2021measuringmassivemultitasklanguage}
Dan Hendrycks, Collin Burns, Steven Basart, Andy Zou, Mantas Mazeika, Dawn
  Song, and Jacob Steinhardt. 2021.
\newblock \href {http://arxiv.org/abs/2009.03300} {Measuring massive multitask
  language understanding}.

\bibitem[{Huang et~al.(2023)Huang, Tang, Zhang, Zhao, Song, Xia, and
  Wei}]{huang2023languagescreatedequalllms}
Haoyang Huang, Tianyi Tang, Dongdong Zhang, Wayne~Xin Zhao, Ting Song, Yan Xia,
  and Furu Wei. 2023.
\newblock \href {http://arxiv.org/abs/2305.07004} {Not all languages are
  created equal in llms: Improving multilingual capability by
  cross-lingual-thought prompting}.

\bibitem[{Hurst et~al.(2024)Hurst, Lerer, Goucher, Perelman, Ramesh, Clark,
  Ostrow, Welihinda, Hayes, Radford et~al.}]{hurst2024gpt}
Aaron Hurst, Adam Lerer, Adam~P Goucher, Adam Perelman, Aditya Ramesh, Aidan
  Clark, AJ~Ostrow, Akila Welihinda, Alan Hayes, Alec Radford, et~al. 2024.
\newblock Gpt-4o system card.
\newblock \emph{arXiv preprint arXiv:2410.21276}.

\bibitem[{Husain et~al.(2024)Husain, Dabre, Kumar, Gala, Jayakumar, Puduppully,
  and Kunchukuttan}]{husain2024romansetuefficientlyunlockingmultilingual}
Jaavid~Aktar Husain, Raj Dabre, Aswanth Kumar, Jay Gala, Thanmay Jayakumar,
  Ratish Puduppully, and Anoop Kunchukuttan. 2024.
\newblock \href {http://arxiv.org/abs/2401.14280} {Romansetu: Efficiently
  unlocking multilingual capabilities of large language models via
  romanization}.

\bibitem[{Indurthi et~al.(2024)Indurthi, Zhou, Chollampatt, Agrawal, Song,
  Zhao, and Zhu}]{indurthi2024improvingmultilingualinstructionfinetuning}
Sathish~Reddy Indurthi, Wenxuan Zhou, Shamil Chollampatt, Ravi Agrawal,
  Kaiqiang Song, Lingxiao Zhao, and Chenguang Zhu. 2024.
\newblock \href {http://arxiv.org/abs/2407.01853} {Improving multilingual
  instruction finetuning via linguistically natural and diverse datasets}.

\bibitem[{Kamath et~al.(2025)Kamath, Ferret, Pathak, Vieillard, Merhej, Perrin,
  Matejovicova, Ram{\'{e}}, Rivi{\`{e}}re, Rouillard, Mesnard, Cideron, Grill,
  Ramos, Yvinec, Casbon, Pot, Penchev, Liu, Visin, Kenealy, Beyer, Zhai,
  Tsitsulin, Busa{-}Fekete, Feng, Sachdeva, Coleman, Gao, Mustafa, Barr,
  Parisotto, Tian, Eyal, Cherry, Peter, Sinopalnikov, Bhupatiraju, Agarwal,
  Kazemi, Malkin, Kumar, Vilar, Brusilovsky, Luo, Steiner, Friesen, Sharma,
  Sharma, Gilady, Goedeckemeyer, Saade, Kolesnikov, Bendebury, Abdagic, Vadi,
  Gy{\"{o}}rgy, Pinto, Das, Bapna, Miech, Yang, Paterson, Shenoy, Chakrabarti,
  Piot, Wu, Shahriari, Petrini, Chen, Lan, Choquette{-}Choo, Carey, Brick,
  Deutsch, Eisenbud, Cattle, Cheng, Paparas, Sreepathihalli, Reid, Tran, Zelle,
  Noland, Huizenga, Kharitonov, Liu, Amirkhanyan, Cameron, Hashemi,
  Klimczak{-}Plucinska, Singh, Mehta, Lehri, Hazimeh, Ballantyne, Szpektor, and
  Nardini}]{gemma3}
Aishwarya Kamath, Johan Ferret, Shreya Pathak, Nino Vieillard, Ramona Merhej,
  Sarah Perrin, Tatiana Matejovicova, Alexandre Ram{\'{e}}, Morgane
  Rivi{\`{e}}re, Louis Rouillard, Thomas Mesnard, Geoffrey Cideron,
  Jean{-}Bastien Grill, Sabela Ramos, Edouard Yvinec, Michelle Casbon, Etienne
  Pot, Ivo Penchev, Ga{\"{e}}l Liu, Francesco Visin, Kathleen Kenealy, Lucas
  Beyer, Xiaohai Zhai, Anton Tsitsulin, R{\'{o}}bert Busa{-}Fekete, Alex Feng,
  Noveen Sachdeva, Benjamin Coleman, Yi~Gao, Basil Mustafa, Iain Barr, Emilio
  Parisotto, David Tian, Matan Eyal, Colin Cherry, Jan{-}Thorsten Peter, Danila
  Sinopalnikov, Surya Bhupatiraju, Rishabh Agarwal, Mehran Kazemi, Dan Malkin,
  Ravin Kumar, David Vilar, Idan Brusilovsky, Jiaming Luo, Andreas Steiner, Abe
  Friesen, Abhanshu Sharma, Abheesht Sharma, Adi~Mayrav Gilady, Adrian
  Goedeckemeyer, Alaa Saade, Alexander Kolesnikov, Alexei Bendebury, Alvin
  Abdagic, Amit Vadi, Andr{\'{a}}s Gy{\"{o}}rgy, Andr{\'{e}}~Susano Pinto, Anil
  Das, Ankur Bapna, Antoine Miech, Antoine Yang, Antonia Paterson, Ashish
  Shenoy, Ayan Chakrabarti, Bilal Piot, Bo~Wu, Bobak Shahriari, Bryce Petrini,
  Charlie Chen, Charline~Le Lan, Christopher~A. Choquette{-}Choo, CJ~Carey,
  Cormac Brick, Daniel Deutsch, Danielle Eisenbud, Dee Cattle, Derek Cheng,
  Dimitris Paparas, Divyashree~Shivakumar Sreepathihalli, Doug Reid, Dustin
  Tran, Dustin Zelle, Eric Noland, Erwin Huizenga, Eugene Kharitonov, Frederick
  Liu, Gagik Amirkhanyan, Glenn Cameron, Hadi Hashemi, Hanna
  Klimczak{-}Plucinska, Harman Singh, Harsh Mehta, Harshal~Tushar Lehri,
  Hussein Hazimeh, Ian Ballantyne, Idan Szpektor, and Ivan Nardini. 2025.
\newblock Gemma 3 technical report.
\newblock \emph{CoRR}, abs/2503.19786.

\bibitem[{Kudugunta et~al.(2023)Kudugunta, Caswell, Zhang, Garcia, Xin,
  Kusupati, Stella, Bapna, and Firat}]{MADLAD}
Sneha Kudugunta, Isaac Caswell, Biao Zhang, Xavier Garcia, Derrick Xin, Aditya
  Kusupati, Romi Stella, Ankur Bapna, and Orhan Firat. 2023.
\newblock {MADLAD-400:} {A} multilingual and document-level large audited
  dataset.
\newblock In \emph{NeurIPS}.

\bibitem[{Kuulmets et~al.(2024)Kuulmets, Purason, Luhtaru, and
  Fishel}]{kuulmets2024teachingllamanewlanguage}
Hele-Andra Kuulmets, Taido Purason, Agnes Luhtaru, and Mark Fishel. 2024.
\newblock \href {http://arxiv.org/abs/2404.04042} {Teaching llama a new
  language through cross-lingual knowledge transfer}.

\bibitem[{Lai and Nissim(2024)}]{lai2024mcotmultilingualinstructiontuning}
Huiyuan Lai and Malvina Nissim. 2024.
\newblock \href {http://arxiv.org/abs/2406.02301} {mcot: Multilingual
  instruction tuning for reasoning consistency in language models}.

\bibitem[{Lai et~al.(2023)Lai, Nguyen, Ngo, Nguyen, Dernoncourt, Rossi, and
  Nguyen}]{lai2023okapiinstructiontunedlargelanguage}
Viet~Dac Lai, Chien~Van Nguyen, Nghia~Trung Ngo, Thuat Nguyen, Franck
  Dernoncourt, Ryan~A. Rossi, and Thien~Huu Nguyen. 2023.
\newblock \href {http://arxiv.org/abs/2307.16039} {Okapi: Instruction-tuned
  large language models in multiple languages with reinforcement learning from
  human feedback}.

\bibitem[{Lample and
  Conneau(2019)}]{lample2019crosslinguallanguagemodelpretraining}
Guillaume Lample and Alexis Conneau. 2019.
\newblock \href {http://arxiv.org/abs/1901.07291} {Cross-lingual language model
  pretraining}.

\bibitem[{Le~Scao et~al.(2023)Le~Scao, Fan, Akiki, Pavlick, Ili{\'c}, Hesslow,
  Castagn{\'e}, Luccioni, Yvon, Gall{\'e} et~al.}]{le2023bloom}
Teven Le~Scao, Angela Fan, Christopher Akiki, Ellie Pavlick, Suzana Ili{\'c},
  Daniel Hesslow, Roman Castagn{\'e}, Alexandra~Sasha Luccioni, Fran{\c{c}}ois
  Yvon, Matthias Gall{\'e}, et~al. 2023.
\newblock Bloom: A 176b-parameter open-access multilingual language model.

\bibitem[{Liu et~al.(2024)Liu, Zhang, Zhao, Luu, and Bing}]{liu2024translation}
Chaoqun Liu, Wenxuan Zhang, Yiran Zhao, Anh~Tuan Luu, and Lidong Bing. 2024.
\newblock Is translation all you need? a study on solving multilingual tasks
  with large language models.
\newblock \emph{arXiv preprint arXiv:2403.10258}.

\bibitem[{Liu et~al.(2025)Liu, Zhang, Zhao, Luu, and
  Bing}]{liu2025translationneedstudysolving}
Chaoqun Liu, Wenxuan Zhang, Yiran Zhao, Anh~Tuan Luu, and Lidong Bing. 2025.
\newblock \href {http://arxiv.org/abs/2403.10258} {Is translation all you need?
  a study on solving multilingual tasks with large language models}.

\bibitem[{LLM-jp et~al.(2024)LLM-jp, :, Aizawa, Aramaki, Chen, Cheng, Deguchi,
  Enomoto, Fujii, Fukumoto, Fukushima, Han, Harada, Hashimoto, Hiraoka, Hisada,
  Hosokawa, Jie, Kamata, Kanazawa, Kanezashi, Kataoka, Katsumata, Kawahara,
  Kawano, Keyaki, Kiryu, Kiyomaru, Kodama, Kubo, Kuga, Kumon, Kurita,
  Kurohashi, Li, Maekawa, Matsuda, Miyao, Mizuki, Mizuki, Murawaki, Mousterou,
  Nakamura, Nakamura, Nakayama, Nakazato, Niitsuma, Nishitoba, Oda, Ogawa,
  Okamoto, Okazaki, Oseki, Ozaki, Ryu, Rzepka, Sakaguchi, Sasaki, Sekine, Suda,
  Sugawara, Sugiura, Sugiyama, Suzuki, Suzuki, Suzumura, Tachibana, Takagi,
  Takami, Takeda, Takeshita, Tanaka, Taura, Tolmachev, Ueda, Wan, Yada, Yahata,
  Yamamoto, Yamauchi, Yanaka, Yokota, and
  Yoshino}]{llmjp2024llmjpcrossorganizationalprojectresearch}
LLM-jp, :, Akiko Aizawa, Eiji Aramaki, Bowen Chen, Fei Cheng, Hiroyuki Deguchi,
  Rintaro Enomoto, Kazuki Fujii, Kensuke Fukumoto, Takuya Fukushima, Namgi Han,
  Yuto Harada, Chikara Hashimoto, Tatsuya Hiraoka, Shohei Hisada, Sosuke
  Hosokawa, Lu~Jie, Keisuke Kamata, Teruhito Kanazawa, Hiroki Kanezashi,
  Hiroshi Kataoka, Satoru Katsumata, Daisuke Kawahara, Seiya Kawano, Atsushi
  Keyaki, Keisuke Kiryu, Hirokazu Kiyomaru, Takashi Kodama, Takahiro Kubo,
  Yohei Kuga, Ryoma Kumon, Shuhei Kurita, Sadao Kurohashi, Conglong Li, Taiki
  Maekawa, Hiroshi Matsuda, Yusuke Miyao, Kentaro Mizuki, Sakae Mizuki, Yugo
  Murawaki, Akim Mousterou, Ryo Nakamura, Taishi Nakamura, Kouta Nakayama,
  Tomoka Nakazato, Takuro Niitsuma, Jiro Nishitoba, Yusuke Oda, Hayato Ogawa,
  Takumi Okamoto, Naoaki Okazaki, Yohei Oseki, Shintaro Ozaki, Koki Ryu, Rafal
  Rzepka, Keisuke Sakaguchi, Shota Sasaki, Satoshi Sekine, Kohei Suda, Saku
  Sugawara, Issa Sugiura, Hiroaki Sugiyama, Hisami Suzuki, Jun Suzuki, Toyotaro
  Suzumura, Kensuke Tachibana, Yu~Takagi, Kyosuke Takami, Koichi Takeda,
  Masashi Takeshita, Masahiro Tanaka, Kenjiro Taura, Arseny Tolmachev, Nobuhiro
  Ueda, Zhen Wan, Shuntaro Yada, Sakiko Yahata, Yuya Yamamoto, Yusuke Yamauchi,
  Hitomi Yanaka, Rio Yokota, and Koichiro Yoshino. 2024.
\newblock \href {http://arxiv.org/abs/2407.03963} {Llm-jp: A
  cross-organizational project for the research and development of fully open
  japanese llms}.

\bibitem[{Muennighoff et~al.(2023)Muennighoff, Wang, Sutawika, Roberts,
  Biderman, Scao, Bari, Shen, Yong, Schoelkopf, Tang, Radev, Aji, Almubarak,
  Albanie, Alyafeai, Webson, Raff, and
  Raffel}]{muennighoff2023crosslingualgeneralizationmultitaskfinetuning}
Niklas Muennighoff, Thomas Wang, Lintang Sutawika, Adam Roberts, Stella
  Biderman, Teven~Le Scao, M~Saiful Bari, Sheng Shen, Zheng-Xin Yong, Hailey
  Schoelkopf, Xiangru Tang, Dragomir Radev, Alham~Fikri Aji, Khalid Almubarak,
  Samuel Albanie, Zaid Alyafeai, Albert Webson, Edward Raff, and Colin Raffel.
  2023.
\newblock \href {http://arxiv.org/abs/2211.01786} {Crosslingual generalization
  through multitask finetuning}.

\bibitem[{Nanda et~al.(2023)Nanda, Chan, Lieberum, Smith, and
  Steinhardt}]{nanda2023progressmeasuresgrokkingmechanistic}
Neel Nanda, Lawrence Chan, Tom Lieberum, Jess Smith, and Jacob Steinhardt.
  2023.
\newblock \href {http://arxiv.org/abs/2301.05217} {Progress measures for
  grokking via mechanistic interpretability}.

\bibitem[{Nguyen et~al.(2024)Nguyen, Nguyen, Lai, Man, Ngo, Dernoncourt, Rossi,
  and Nguyen}]{CulturaX}
Thuat Nguyen, Chien~Van Nguyen, Viet~Dac Lai, Hieu Man, Nghia~Trung Ngo, Franck
  Dernoncourt, Ryan~A. Rossi, and Thien~Huu Nguyen. 2024.
\newblock Culturax: {A} cleaned, enormous, and multilingual dataset for large
  language models in 167 languages.
\newblock In \emph{{LREC/COLING}}, pages 4226--4237. {ELRA} and {ICCL}.

\bibitem[{Ni et~al.(2023)Ni, Mao, Yang, Lei, and Cambria}]{ni2023finding}
Jinjie Ni, Rui Mao, Zonglin Yang, Han Lei, and Erik Cambria. 2023.
\newblock Finding the pillars of strength for multi-head attention.
\newblock In \emph{Proceedings of the 61st Annual Meeting of the Association
  for Computational Linguistics (Volume 1: Long Papers)}, pages 14526--14540.

\bibitem[{Nooralahzadeh et~al.(2020)Nooralahzadeh, Bekoulis, Bjerva, and
  Augenstein}]{nooralahzadeh-etal-2020-zero}
Farhad Nooralahzadeh, Giannis Bekoulis, Johannes Bjerva, and Isabelle
  Augenstein. 2020.
\newblock \href {https://doi.org/10.18653/v1/2020.emnlp-main.368} {Zero-shot
  cross-lingual transfer with meta learning}.
\newblock In \emph{Proceedings of the 2020 Conference on Empirical Methods in
  Natural Language Processing (EMNLP)}, pages 4547--4562, Online. Association
  for Computational Linguistics.

\bibitem[{OpenAI(2023)}]{openai2023gpt4}
OpenAI. 2023.
\newblock \href {http://arxiv.org/abs/2303.08774} {Gpt-4 technical report}.

\bibitem[{OpenAI(2024)}]{openai2024mmmlu}
OpenAI. 2024.
\newblock \href {https://huggingface.co/datasets/openai/MMMLU} {Multilingual
  massive multitask language understanding (mmmlu)}.
\newblock Hugging Face.

\bibitem[{Pires et~al.(2019)Pires, Schlinger, and
  Garrette}]{pires-etal-2019-multilingual}
Telmo Pires, Eva Schlinger, and Dan Garrette. 2019.
\newblock \href {https://doi.org/10.18653/v1/P19-1493} {How multilingual is
  multilingual {BERT}?}
\newblock In \emph{Proceedings of the 57th Annual Meeting of the Association
  for Computational Linguistics}, pages 4996--5001, Florence, Italy.
  Association for Computational Linguistics.

\bibitem[{Qin et~al.(2023)Qin, Chen, Wei, Huang, and Che}]{qin2023cross}
Libo Qin, Qiguang Chen, Fuxuan Wei, Shijue Huang, and Wanxiang Che. 2023.
\newblock Cross-lingual prompting: Improving zero-shot chain-of-thought
  reasoning across languages.
\newblock In \emph{The 2023 Conference on Empirical Methods in Natural Language
  Processing}.

\bibitem[{Schut et~al.(2025{\natexlab{a}})Schut, Gal, and
  Farquhar}]{schut2025multilingual}
Lisa Schut, Yarin Gal, and Sebastian Farquhar. 2025{\natexlab{a}}.
\newblock Do multilingual llms think in english?
\newblock \emph{arXiv preprint arXiv:2502.15603}.

\bibitem[{Schut et~al.(2025{\natexlab{b}})Schut, Gal, and
  Farquhar}]{schut2025multilingualllmsthinkenglish}
Lisa Schut, Yarin Gal, and Sebastian Farquhar. 2025{\natexlab{b}}.
\newblock \href {http://arxiv.org/abs/2502.15603} {Do multilingual llms think
  in english?}

\bibitem[{Shi et~al.(2022)Shi, Suzgun, Freitag, Wang, Srivats, Vosoughi, Chung,
  Tay, Ruder, Zhou et~al.}]{shi2022language}
Freda Shi, Mirac Suzgun, Markus Freitag, Xuezhi Wang, Suraj Srivats, Soroush
  Vosoughi, Hyung~Won Chung, Yi~Tay, Sebastian Ruder, Denny Zhou, et~al. 2022.
\newblock Language models are multilingual chain-of-thought reasoners.
\newblock In \emph{The Eleventh International Conference on Learning
  Representations}.

\bibitem[{Stanczak et~al.(2022)Stanczak, Ponti, Torroba~Hennigen, Cotterell,
  and Augenstein}]{stanczak-etal-2022-neurons}
Karolina Stanczak, Edoardo Ponti, Lucas Torroba~Hennigen, Ryan Cotterell, and
  Isabelle Augenstein. 2022.
\newblock \href {https://doi.org/10.18653/v1/2022.naacl-main.114} {Same
  neurons, different languages: Probing morphosyntax in multilingual
  pre-trained models}.
\newblock In \emph{Proceedings of the 2022 Conference of the North American
  Chapter of the Association for Computational Linguistics: Human Language
  Technologies}, pages 1589--1598, Seattle, United States. Association for
  Computational Linguistics.

\bibitem[{Tang et~al.(2024)Tang, Luo, Huang, Zhang, Wang, Zhao, Wei, and
  Wen}]{tang2024language}
Tianyi Tang, Wenyang Luo, Haoyang Huang, Dongdong Zhang, Xiaolei Wang,
  Wayne~Xin Zhao, Furu Wei, and Ji-Rong Wen. 2024.
\newblock Language-specific neurons: The key to multilingual capabilities in
  large language models.
\newblock In \emph{Proceedings of the 62nd Annual Meeting of the Association
  for Computational Linguistics (Volume 1: Long Papers)}, pages 5701--5715.

\bibitem[{Team et~al.(2024)Team, Georgiev, Lei, Burnell, Bai, Gulati, Tanzer,
  Vincent, Pan, Wang et~al.}]{team2024gemini}
Gemini Team, Petko Georgiev, Ving~Ian Lei, Ryan Burnell, Libin Bai, Anmol
  Gulati, Garrett Tanzer, Damien Vincent, Zhufeng Pan, Shibo Wang, et~al. 2024.
\newblock Gemini 1.5: Unlocking multimodal understanding across millions of
  tokens of context.
\newblock \emph{arXiv preprint arXiv:2403.05530}.

\bibitem[{Touvron et~al.(2023{\natexlab{a}})Touvron, Lavril, Izacard, Martinet,
  Lachaux, Lacroix, Rozi{\`e}re, Goyal, Hambro, Azhar
  et~al.}]{touvron2023llama}
Hugo Touvron, Thibaut Lavril, Gautier Izacard, Xavier Martinet, Marie-Anne
  Lachaux, Timoth{\'e}e Lacroix, Baptiste Rozi{\`e}re, Naman Goyal, Eric
  Hambro, Faisal Azhar, et~al. 2023{\natexlab{a}}.
\newblock Llama: Open and efficient foundation language models.
\newblock \emph{arXiv preprint arXiv:2302.13971}.

\bibitem[{Touvron et~al.(2023{\natexlab{b}})Touvron, Martin, Stone, Albert,
  Almahairi, Babaei, Bashlykov, Batra, Bhargava, Bhosale
  et~al.}]{touvron2023llama2}
Hugo Touvron, Louis Martin, Kevin Stone, Peter Albert, Amjad Almahairi, Yasmine
  Babaei, Nikolay Bashlykov, Soumya Batra, Prajjwal Bhargava, Shruti Bhosale,
  et~al. 2023{\natexlab{b}}.
\newblock Llama 2: Open foundation and fine-tuned chat models.
\newblock \emph{arXiv preprint arXiv:2307.09288}.

\bibitem[{{\"U}st{\"u}n et~al.(2024){\"U}st{\"u}n, Aryabumi, Yong, Ko, D'souza,
  Onilude, Bhandari, Singh, Ooi, Kayid et~al.}]{ustun2024aya}
Ahmet {\"U}st{\"u}n, Viraat Aryabumi, Zheng-Xin Yong, Wei-Yin Ko, Daniel
  D'souza, Gbemileke Onilude, Neel Bhandari, Shivalika Singh, Hui-Lee Ooi, Amr
  Kayid, et~al. 2024.
\newblock Aya model: An instruction finetuned open-access multilingual language
  model.
\newblock \emph{arXiv preprint arXiv:2402.07827}.

\bibitem[{Wang et~al.(2025{\natexlab{a}})Wang, Zhao, Cai, and
  Tan}]{wang2025investigating}
Chengxin Wang, Yiran Zhao, Shaofeng Cai, and Gary Tan. 2025{\natexlab{a}}.
\newblock \href {https://openreview.net/forum?id=a9vey6B54y} {Investigating
  pattern neurons in urban time series forecasting}.
\newblock In \emph{The Thirteenth International Conference on Learning
  Representations}.

\bibitem[{Wang et~al.(2025{\natexlab{b}})Wang, He, Yu, Fu, and
  Han}]{wang2025largelanguagemodelsgood}
Teng Wang, Zhenqi He, Wing-Yin Yu, Xiaojin Fu, and Xiongwei Han.
  2025{\natexlab{b}}.
\newblock \href {http://arxiv.org/abs/2409.11056} {Large language models are
  good multi-lingual learners : When llms meet cross-lingual prompts}.

\bibitem[{Wendler et~al.(2024)Wendler, Veselovsky, Monea, and
  West}]{wendler2024llamas}
Chris Wendler, Veniamin Veselovsky, Giovanni Monea, and Robert West. 2024.
\newblock Do llamas work in english? on the latent language of multilingual
  transformers.
\newblock In \emph{Proceedings of the 62nd Annual Meeting of the Association
  for Computational Linguistics (Volume 1: Long Papers)}, pages 15366--15394.

\bibitem[{Wu and Dredze(2019)}]{wu-dredze-2019-beto}
Shijie Wu and Mark Dredze. 2019.
\newblock \href {https://doi.org/10.18653/v1/D19-1077} {Beto, bentz, becas: The
  surprising cross-lingual effectiveness of {BERT}}.
\newblock In \emph{Proceedings of the 2019 Conference on Empirical Methods in
  Natural Language Processing and the 9th International Joint Conference on
  Natural Language Processing (EMNLP-IJCNLP)}, pages 833--844, Hong Kong,
  China. Association for Computational Linguistics.

\bibitem[{Wu et~al.(2025)Wu, Yu, Yogatama, Lu, and
  Kim}]{wu2025semantichubhypothesislanguage}
Zhaofeng Wu, Xinyan~Velocity Yu, Dani Yogatama, Jiasen Lu, and Yoon Kim. 2025.
\newblock \href {http://arxiv.org/abs/2411.04986} {The semantic hub hypothesis:
  Language models share semantic representations across languages and
  modalities}.

\bibitem[{Yang et~al.(2024{\natexlab{a}})Yang, Yang, Hui, Zheng, Yu, Zhou, Li,
  Li, Liu, Huang, Dong, Wei, Lin, Tang, Wang, Yang, Tu, Zhang, Ma, Yang, Xu,
  Zhou, Bai, He, Lin, Dang, Lu, Chen, Yang, Li, Xue, Ni, Zhang, Wang, Peng,
  Men, Gao, Lin, Wang, Bai, Tan, Zhu, Li, Liu, Ge, Deng, Zhou, Ren, Zhang, Wei,
  Ren, Liu, Fan, Yao, Zhang, Wan, Chu, Liu, Cui, Zhang, Guo, and
  Fan}]{yang2024qwen2technicalreport}
An~Yang, Baosong Yang, Binyuan Hui, Bo~Zheng, Bowen Yu, Chang Zhou, Chengpeng
  Li, Chengyuan Li, Dayiheng Liu, Fei Huang, Guanting Dong, Haoran Wei, Huan
  Lin, Jialong Tang, Jialin Wang, Jian Yang, Jianhong Tu, Jianwei Zhang,
  Jianxin Ma, Jianxin Yang, Jin Xu, Jingren Zhou, Jinze Bai, Jinzheng He,
  Junyang Lin, Kai Dang, Keming Lu, Keqin Chen, Kexin Yang, Mei Li, Mingfeng
  Xue, Na~Ni, Pei Zhang, Peng Wang, Ru~Peng, Rui Men, Ruize Gao, Runji Lin,
  Shijie Wang, Shuai Bai, Sinan Tan, Tianhang Zhu, Tianhao Li, Tianyu Liu,
  Wenbin Ge, Xiaodong Deng, Xiaohuan Zhou, Xingzhang Ren, Xinyu Zhang, Xipin
  Wei, Xuancheng Ren, Xuejing Liu, Yang Fan, Yang Yao, Yichang Zhang, Yu~Wan,
  Yunfei Chu, Yuqiong Liu, Zeyu Cui, Zhenru Zhang, Zhifang Guo, and Zhihao Fan.
  2024{\natexlab{a}}.
\newblock \href {http://arxiv.org/abs/2407.10671} {Qwen2 technical report}.

\bibitem[{Yang et~al.(2024{\natexlab{b}})Yang, Yang, Zhang, Hui, Zheng, Yu, Li,
  Liu, Huang, Wei et~al.}]{yang2024qwen2}
An~Yang, Baosong Yang, Beichen Zhang, Binyuan Hui, Bo~Zheng, Bowen Yu,
  Chengyuan Li, Dayiheng Liu, Fei Huang, Haoran Wei, et~al. 2024{\natexlab{b}}.
\newblock Qwen2. 5 technical report.
\newblock \emph{arXiv preprint arXiv:2412.15115}.

\bibitem[{Zhang et~al.(2024)Zhang, Chen, Li, Che, and
  Qin}]{zhang2024autocapautomaticcrosslingualalignment}
Yongheng Zhang, Qiguang Chen, Min Li, Wanxiang Che, and Libo Qin. 2024.
\newblock \href {http://arxiv.org/abs/2406.13940} {Autocap: Towards automatic
  cross-lingual alignment planning for zero-shot chain-of-thought}.

\bibitem[{Zhang et~al.(2021)Zhang, Gu, Han, Chen, Xiao, Sun, Yao, Qi, Guan, Ke,
  Cai, Zeng, Tan, Liu, Huang, Han, Liu, Zhu, and
  Sun}]{zhang2021cpm2largescalecosteffectivepretrained}
Zhengyan Zhang, Yuxian Gu, Xu~Han, Shengqi Chen, Chaojun Xiao, Zhenbo Sun, Yuan
  Yao, Fanchao Qi, Jian Guan, Pei Ke, Yanzheng Cai, Guoyang Zeng, Zhixing Tan,
  Zhiyuan Liu, Minlie Huang, Wentao Han, Yang Liu, Xiaoyan Zhu, and Maosong
  Sun. 2021.
\newblock \href {http://arxiv.org/abs/2106.10715} {Cpm-2: Large-scale
  cost-effective pre-trained language models}.

\bibitem[{Zhao et~al.(2025)Zhao, Liu, Deng, Ying, Aljunied, Li, Bing, Chan,
  Rong, Zhao et~al.}]{zhao2025babel}
Yiran Zhao, Chaoqun Liu, Yue Deng, Jiahao Ying, Mahani Aljunied, Zhaodonghui
  Li, Lidong Bing, Hou~Pong Chan, Yu~Rong, Deli Zhao, et~al. 2025.
\newblock Babel: Open multilingual large language models serving over 90\% of
  global speakers.
\newblock \emph{arXiv preprint arXiv:2503.00865}.

\bibitem[{Zhao et~al.(2024{\natexlab{a}})Zhao, Zhang, Chen, Kawaguchi, and
  Bing}]{how_do_handle}
Yiran Zhao, Wenxuan Zhang, Guizhen Chen, Kenji Kawaguchi, and Lidong Bing.
  2024{\natexlab{a}}.
\newblock How do large language models handle multilingualism?
\newblock In \emph{NeurIPS}.

\bibitem[{Zhao et~al.(2024{\natexlab{b}})Zhao, Zhang, Wang, Kawaguchi, and
  Bing}]{zhao2024adamergexcrosslingualtransferlarge}
Yiran Zhao, Wenxuan Zhang, Huiming Wang, Kenji Kawaguchi, and Lidong Bing.
  2024{\natexlab{b}}.
\newblock \href {http://arxiv.org/abs/2402.18913} {Adamergex: Cross-lingual
  transfer with large language models via adaptive adapter merging}.

\bibitem[{Zhong et~al.(2024)Zhong, Cheng, Liu, Jiang, Wan, Chu, Murawaki, and
  Kurohashi}]{zhong2024englishcentricllmslanguagemultilingual}
Chengzhi Zhong, Fei Cheng, Qianying Liu, Junfeng Jiang, Zhen Wan, Chenhui Chu,
  Yugo Murawaki, and Sadao Kurohashi. 2024.
\newblock \href {http://arxiv.org/abs/2408.10811} {Beyond english-centric llms:
  What language do multilingual language models think in?}

\end{thebibliography}




\clearpage

\appendix

\section{Parallel Neuron Detection Algorithm}\label{sec:parallel}

Inspired by \citet{how_do_handle, wang2025investigating}, the neuron detection method in Equation \ref{equ:neuron} can be done parallel. 
While Equation~\ref{equ:neuron} considers the change in the final output embedding, the parallel methods described here efficiently calculate the change in the output of the \textit{specific layer containing the neuron} when that neuron is deactivated. This layer-wise impact serves as a proxy or component for the overall impact.

In this context, let $X \in \mathbb{R}^{l \times d_{model}}$ be the input hidden states to a given layer, where $l$ is the sequence length and $d_{model}$ is the hidden dimension of the model. For a neuron $\mathcal{N}$ within this layer, its impact is measured as $\| f(X; \Theta) - f(X; \Theta_{\ominus \mathcal{N}}) \|_2$, where $f(X; \Theta)$ is the layer's output with parameters $\Theta$, and $f(X; \Theta_{\ominus \mathcal{N}})$ is the output when neuron $\mathcal{N}$ (a specific row or column in $\Theta$) is deactivated (its parameters set to zero).

\subsection{Feed-Forward Network (FFN) Neurons}

A standard FFN layer in modern transformer models can be expressed as:
\begin{equation}
\text{FFN}(X) = \left(\text{SiLU}(XW_{gate}) \odot (XW_{up})\right)W_{down}
\end{equation}
where $X \in \mathbb{R}^{l \times d_{model}}$ is the input to the FFN layer, $W_{gate}, W_{up} \in \mathbb{R}^{d_{model} \times d_{inter}}$, and $W_{down} \in \mathbb{R}^{d_{inter} \times d_{model}}$. Here, $d_{inter}$ is the intermediate dimension of the FFN. The symbol $\odot$ denotes element-wise multiplication. Let $H_{act} = \text{SiLU}(XW_{gate}) \odot (XW_{up})$ be the intermediate activation matrix, $H_{act} \in \mathbb{R}^{l \times d_{inter}}$. Thus, the FFN output is $Y_{FFN} = H_{act}W_{down} \in \mathbb{R}^{l \times d_{model}}$.

We consider a neuron $\mathcal{N}_{inter,k}$ to be associated with the $k$-th dimension of the intermediate representation $H_{act}$. Deactivating such a neuron means that the $k$-th column of $H_{act}$, denoted $H_{act}[:,k]$, is effectively zeroed out before the multiplication with $W_{down}$. This deactivation corresponds to zeroing out the parameters that produce this $k$-th intermediate feature, e.g., the $k$-th column of $W_{up}$ (i.e., neuron $\mathcal{N}$ is $W_{up}[:,k]$) and $W_{gate}$, or by zeroing out parameters that read from it, e.g., the $k$-th row of $W_{down}$ (i.e., neuron $\mathcal{N}$ is $W_{down}[k,:]$).

Let $Y_{FFN,\ominus \mathcal{N}_{inter,k}}$ be the output when the $k$-th intermediate neuron is deactivated. The change in the layer's output is:
\[
\Delta Y_{FFN,k} = Y_{FFN} - Y_{FFN,\ominus \mathcal{N}_{inter,k}}
\]
If $H'_{act}$ is $H_{act}$ with its $k$-th column zeroed, then $Y_{FFN,\ominus \mathcal{N}_{inter,k}} = H'_{act}W_{down}$.
So,
\[
\Delta Y_{FFN,k} = (H_{act} - H'_{act})W_{down}
\]
The matrix $(H_{act} - H'_{act})$ is zero everywhere except for its $k$-th column, which consists of the elements $H_{act}[:,k]$. Let this difference matrix be $\delta H_k$.
Then $\Delta Y_{FFN,k} = \delta H_k W_{down}$. This resulting $l \times d_{model}$ matrix is formed by the outer product of the $k$-th column of $H_{act}$ and the $k$-th row of $W_{down}$:
\[
\Delta Y_{FFN,k} = H_{act}[:,k] (W_{down})_{k,:}
\]
The impact of the $k$-th intermediate FFN neuron is then the L2 norm of this change:
\begin{equation} \label{eq:ffn_impact}
\left\| \Delta Y_{FFN,k} \right\|_2 = \left\| H_{act}[:,k] (W_{down})_{k,:} \right\|_2
\end{equation}
This computation can be performed in parallel for all $k \in \{1, \dots, d_{inter}\}$ to obtain the impact of all intermediate neurons in the FFN layer.

\subsection{Self-Attention Network Neurons}

The output of a self-attention layer (for simplicity, we describe a single attention head; multi-head attention involves similar computations per head) can be given by:
\begin{equation}
Y_{Attn} = \text{Softmax}\left(\frac{(XW_Q)(XW_K)^T}{\sqrt{d_k}}\right)(XW_V)
\end{equation}
Let $Q = XW_Q \in \mathbb{R}^{l \times d_{attn}}$, $K = XW_K \in \mathbb{R}^{l \times d_{attn}}$, and $V = XW_V \in \mathbb{R}^{l \times d_{attn}}$, where $d_{attn}$ is the dimension of queries, keys, and values for the attention mechanism. $d_k$ is the scaling factor, typically the dimension of the key/query vectors (e.g., $d_k = d_{attn}$).
Let $A = \text{Softmax}\left(\frac{QK^T}{\sqrt{d_k}}\right) \in \mathbb{R}^{l \times l}$. The layer output is $Y_{Attn} = AV \in \mathbb{R}^{l \times d_{attn}}$. (An additional output projection $W_O$ might follow this, which would be multiplied subsequently).

\subsubsection{Neurons in $W_V$}
Consider a neuron $\mathcal{N}_{V,k}$ defined as the $k$-th column of $W_V$, i.e., $W_V[:,k]$. Deactivating this neuron sets $W_V[:,k]$ to zero, which in turn makes the $k$-th column of $V = XW_V$, denoted $V[:,k]$, zero.
Let $V'$ be the matrix $V$ with its $k$-th column zeroed. The change in the layer's output is:
\[
\Delta Y_{Attn,k}^{(V)} = AV - AV' = A(V-V')
\]
The matrix $(V-V')$ is zero everywhere except for its $k$-th column, which is $V[:,k]$. Let this difference matrix be $\delta V_k$.
Then $\Delta Y_{Attn,k}^{(V)} = A (\delta V_k)$. This $l \times d_{attn}$ matrix has $A V[:,k]$ (the matrix $A$ multiplied by the vector $V[:,k]$) as its $k$-th column, and zeros in other columns.
The impact of neuron $\mathcal{N}_{V,k}$ is:
\begin{equation} \label{eq:wv_impact}
\left\| \Delta Y_{Attn,k}^{(V)} \right\|_2 = \left\| A V[:,k] \right\|_2
\end{equation}
where the norm is effectively taken over the $l \times 1$ vector $A V[:,k]$ that forms the $k$-th column of the change matrix. This can be calculated in parallel for all $k \in \{1, \dots, d_{attn}\}$.

\subsubsection{Neurons in $W_Q$}
Consider a neuron $\mathcal{N}_{Q,k}$ defined as the $k$-th column of $W_Q$, i.e., $W_Q[:,k]$. Deactivating this neuron sets $W_Q[:,k]$ to zero. This makes the $k$-th column of $Q = XW_Q$, denoted $Q[:,k]$, zero.
Let $Q'$ be the matrix $Q$ with its $k$-th column zeroed.
The original unnormalized attention scores are $S_{raw} = \frac{QK^T}{\sqrt{d_k}}$.
The new unnormalized attention scores with $\mathcal{N}_{Q,k}$ deactivated are $S'_{raw} = \frac{Q'K^T}{\sqrt{d_k}}$.
The change in the unnormalized scores due to deactivating $\mathcal{N}_{Q,k}$ is $\Delta S_{raw, k} = S_{raw} - S'_{raw} = \frac{(Q-Q')K^T}{\sqrt{d_k}}$.
The matrix $(Q-Q')$ is zero everywhere except for its $k$-th column, which is $Q[:,k]$.
Thus,
\[
\Delta S_{raw, k} = \frac{(Q[:,k]) (K[:,k])^T}{\sqrt{d_k}}
\]
This $l \times l$ matrix represents the change in raw attention scores attributable to the interaction involving the $k$-th column of $Q$ and the $k$-th column of $K$.

Let $A_{orig} = \text{softmax}(S_{raw})$ be the original attention probability matrix.
Let $A_{\ominus \mathcal{N}_{Q,k}} = \text{softmax}(S_{raw} - \Delta S_{raw, k})$ be the attention probability matrix when neuron $\mathcal{N}_{Q,k}$ is deactivated.
The change in the layer's output is:
\[
\Delta Y_{Attn,k}^{(Q)} = A_{orig}V - A_{\ominus \mathcal{N}_{Q,k}}V = (A_{orig} - A_{\ominus \mathcal{N}_{Q,k}})V
\]
The impact of neuron $\mathcal{N}_{Q,k}$ is:
\begin{equation} \label{eq:wq_impact}
\left\| \Delta Y_{Attn,k}^{(Q)} \right\|_2 = \left\| (A_{orig} - A_{\ominus \mathcal{N}_{Q,k}})V \right\|_2
\end{equation}
To calculate this efficiently for all $k \in \{1, \dots, d_{attn}\}$ (corresponding to each column neuron in $W_Q$):
\begin{enumerate}
    \item Compute the original $S_{raw} = \frac{QK^T}{\sqrt{d_k}}$ and $A_{orig} = \text{softmax}(S_{raw})$.
    \item For each $k$, compute the specific change term $\Delta S_{raw, k} = \frac{Q[:,k] (K[:,k])^T}{\sqrt{d_k}}$. This step can be parallelized by constructing a tensor ${\Delta S}_{raw} \in \mathbb{R}^{d_{attn} \times l \times l}$ where the slice ${\Delta S}_{raw}[k,:,:] = \Delta S_{raw, k}$.
    \item For each $k$, compute the adjusted scores $S_{adjusted, k} = S_{raw} - {\Delta S}_{raw}[k,:,:]$.
    \item For each $k$, compute $A_{\ominus \mathcal{N}_{Q,k}} = \text{softmax}(S_{adjusted, k})$.
    \item For each $k$, calculate the impact norm $\| (A_{orig} - A_{\ominus \mathcal{N}_{Q,k}})V \|_2$.
\end{enumerate}

\subsubsection{Neurons in $W_K$}
The impact of deactivating a neuron $\mathcal{N}_{K,k}$ (the $k$-th column of $W_K$) is calculated symmetrically to that of $\mathcal{N}_{Q,k}$. The same change term $\Delta S_{raw, k} = \frac{Q[:,k] (K[:,k])^T}{\sqrt{d_k}}$ is used, reflecting the idea that this term captures the interaction component associated with the $k$-th features of both $Q$ and $K$. The procedure then follows steps 3-5 as outlined for $W_Q$ neurons, using this $\Delta S_{raw, k}$ to find the adjusted attention matrix and the resulting impact.

\section{Neuron Detection Corpus} \label{sec:corpus}

This section provides additional details regarding the corpus used for neuron detection, as mentioned in the main text. Our methodology relies on the OSCAR corpus for both identifying language-related neurons through activation patterns and quantifying their functional contribution via perplexity changes upon deactivation.

\subsection{OSCAR Corpus}

The OSCAR (Open Super-large Crawled Aggregated coRpus) corpus~\citep{oscar} is a massive multilingual collection of texts obtained by language classification and filtering of the Common Crawl dataset. Common Crawl is a publicly available web crawl spanning petabytes of data. OSCAR further processes this raw data to produce monolingual corpora across a wide range of languages, making it a valuable resource for training large language models and conducting cross-lingual research.

Key characteristics of the OSCAR corpus include:
\begin{itemize}[leftmargin=*]
    \item \textbf{Large Scale:} It contains hundreds of gigabytes to terabytes of text data for many languages.
    \item \textbf{Multilingual Coverage:} It supports a vast number of languages, facilitating studies that require diverse linguistic data.
    \item \textbf{Data Cleaning:} Efforts are made to clean and filter the crawled data, though the quality can vary depending on the language and the nature of web content.
    \item \textbf{Accessibility:} OSCAR is publicly available, promoting reproducibility and broader research in NLP.
\end{itemize}
For our study, we sample $1000$ sentences for each target language from its respective monolingual section within the OSCAR corpus. This sampled data serves as the basis for analyzing neuron activations and evaluating perplexity changes. The diversity and scale of OSCAR help in capturing a wide array of linguistic phenomena necessary for robustly identifying language-specific neural correlates.

\subsection{Illustration of Sample Sentences}

To provide a concrete illustration of the data used, Table~\ref{tab:corpus_samples} presents conceptual example sentences from the OSCAR corpus for the five languages central to our analysis: English (en), Chinese (zh), Swahili (sw), German (de), and French (fr). 

\begin{table}[h!]
\centering
\caption{Illustrative sample sentences from the OSCAR corpus for the selected languages. These are conceptual examples, as actual sentences are randomly sampled.}
\label{tab:corpus_samples}
\begin{tabular}{ll}
\toprule
\textbf{Language} & \textbf{Conceptual Example} \\
\midrule
English (en) & The quick brown fox jumps over the lazy dog. \\
Chinese (zh) & \zh{敏捷的棕色狐狸跳过了懒惰的狗。} \\
Swahili (sw) & Mbweha mwepesi wa kahawia anaruka juu ya mbwa mvivu. \\
German (de) & Der schnelle braune Fuchs springt über den faulen Hund. \\
French (fr) & Le renard brun rapide saute par-dessus le chien paresseux. \\
\bottomrule
\end{tabular}
\end{table}

The sentences sampled for each language are then further used to observe which neurons are consistently activated during processing. A similar set of sentences is then used to measure the perplexity of the model when specific neurons or sets of neurons are deactivated, thereby quantifying their functional importance to that language.

\end{document}